\documentclass[11pt]{article}

\usepackage[final]{acl}

\usepackage{times}
\usepackage{latexsym}
\usepackage{amsmath}

\usepackage{times}
\usepackage{latexsym}
\usepackage[T1]{fontenc}
\usepackage[utf8]{inputenc}
\usepackage{microtype}
\usepackage{inconsolata}
\usepackage{graphicx}
\usepackage{booktabs}
\usepackage{multirow}
\usepackage{amsmath}
\usepackage{amssymb}
\usepackage{algorithm}
\usepackage{algpseudocode}
\usepackage{url}
\usepackage{xcolor}
\usepackage[table]{xcolor}
\usepackage{subcaption}
\definecolor{tableaublue}{HTML}{1f77b4}

\usepackage[T1]{fontenc}

\usepackage[utf8]{inputenc}

\usepackage{microtype}

\usepackage{inconsolata}

\usepackage{graphicx}

\newcommand*{\img}[1]{%
    \raisebox{-.01\baselineskip}{%
        \includegraphics[
        height=2\baselineskip,
        width=2\baselineskip,
        keepaspectratio,
        ]{#1}%
    }%
}

\usepackage{listings}
\definecolor{diffadd}{HTML}{22863A}
\definecolor{diffrem}{HTML}{B31D28}
\definecolor{diffhunk}{HTML}{6A737D}
\definecolor{diffbg}{HTML}{F6F8FA}
 
\lstdefinelanguage{gitdiff}{
  basicstyle=\ttfamily\scriptsize,
  backgroundcolor=\color{diffbg},
  numbers=none,
  showspaces=false,
  showstringspaces=false,
  showtabs=false,
  breaklines=true,
  breakatwhitespace=true,
  columns=fullflexible,
  keepspaces=true,
  frame=single,
  framerule=0pt,
  framesep=3pt,
  xleftmargin=4pt,
  xrightmargin=4pt,
  morecomment=[l][\color{diffrem}]{-},
  morecomment=[l][\color{diffadd}]{+},
  morecomment=[l][\color{diffhunk}\itshape]{@@},
}

\title{\img{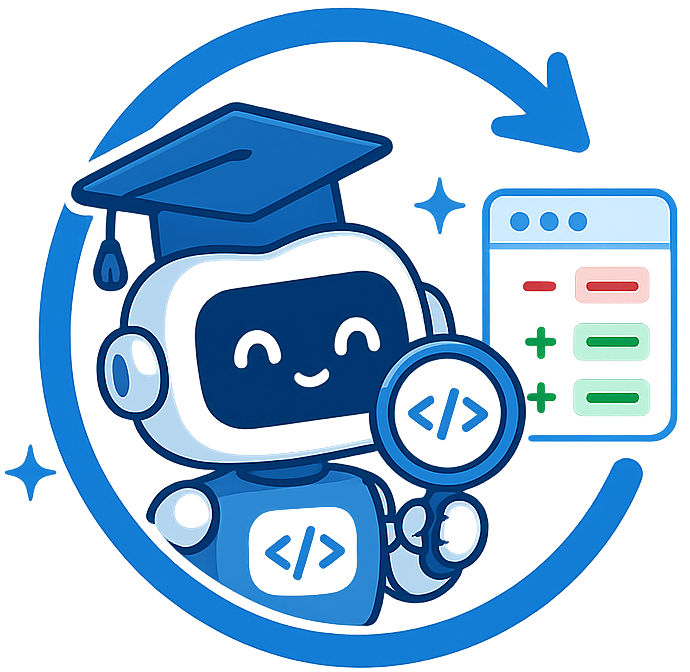} Beneath the Diff: Diagnosing and Mitigating Algorithmic Mode Collapse in Code-Level Autonomous Research Loops}

\author{Bowei He\textsuperscript{1, 2}, Weixu Zhang\textsuperscript{2}, Yili Jin\textsuperscript{3, 4}~\thanks{Corresponding to Dr. Yili Jin.}, Xue Liu\textsuperscript{1, 2} \\
\textsuperscript{1} MBZUAI, \textsuperscript{2} McGill University, \textsuperscript{3} MirrorSpace Technology,
\textsuperscript{4} Simon Fraser University\\
  \texttt{Bowei.He@mbzuai.ac.ae, yili@mirrorspace.tech}}

\usepackage{xargs}
\usepackage[colorinlistoftodos,prependcaption,textsize=tiny]{todonotes} 
\newcommandx{\jz}[2][1=]{\todo[linecolor=magenta,backgroundcolor=magenta!25,bordercolor=magenta,#1]{Jieyu: #2}}

\begin{document}
\maketitle
\begin{abstract}
Code-level autonomous research loops (ARLs) have recently emerged as a concrete object of study in automated machine learning research. In such loops, an LLM agent proposes modifications to an experimental training pipeline, executes the modified pipeline, and retains edits that improve a verifiable in-loop metric. Although executable metrics may appear to provide a reliable signal of progress, it remains unclear whether repeated metric-driven code editing leads to genuine improvements that generalize beyond the loop.
We provide a systematic diagnosis of this question. Across various experiment settings, we identify a robust failure mode that we call \textbf{algorithmic mode collapse}. In this regime, surface-level edit diversity remains stable, but semantic and mechanism-level diversity collapse: the agent continues to edit different lines of code while repeatedly proposing the same kinds of algorithmic changes. This collapse is accompanied by a widening gap between in-loop metric gains and gains measured on independent held-out evaluations. We then propose Diversity-Aware Proposal Sampling (\textsc{DAPS}), a lightweight mitigation that combines category-coverage reweighting, persistent edit memory, and a validation gate. Under a three-tier protocol separating the in-loop metric, the audit metric read by the gate, and a blind metric no loop component ever accesses, \textsc{DAPS} reduces semantic-cluster decay of edits by $69.1\%$ and improves relative faithfulness by $83.7\%$ blind and $81.6\%$ audited, while preserving in-loop optimization speed. We provide the code in Github \href{https://github.com/BokwaiHo/arl-mode-collapse}{repository}.
\end{abstract}

\section{Introduction}
\label{sec:intro}
The recent open-source release of iterative autonomous research agents, like the most prominently \texttt{autoresearch} by \citet{karpathy2026autoresearch} and the broader AI-Scientist family \citep{lu2024aiscientist,schmidgall2025agentlab,airesearcher2025,gottweis2025coscientist}, has made \emph{code-level autonomous research loops} (ARLs) a concrete object of study in automated machine learning research. In such loops, an LLM proposer generates modifications to a training pipeline (optimizer settings, architectural details, data preprocessing, or loss formulations), the modified pipeline is executed, and a validation metric decides whether the edit is retained or reverted. Thus, we can automate the iterative search for better training recipes with minimal human intervention. Unlike recursive training~\citep{recursiveloops2025} which relies on self-generated training samples to improve the model, code-level ARLs can access optimization signals grounded in an external, executable evaluation. Therefore, one might expect autonomous research loops to be immune to the diversity collapse in recursive training loops, which narrows the space of generated solutions and undermines generalization \citep{shumailov2024curse,dces2026,prism2026}.
 
\begin{figure*}[t]
  \centering
  \includegraphics[width=0.85\linewidth]{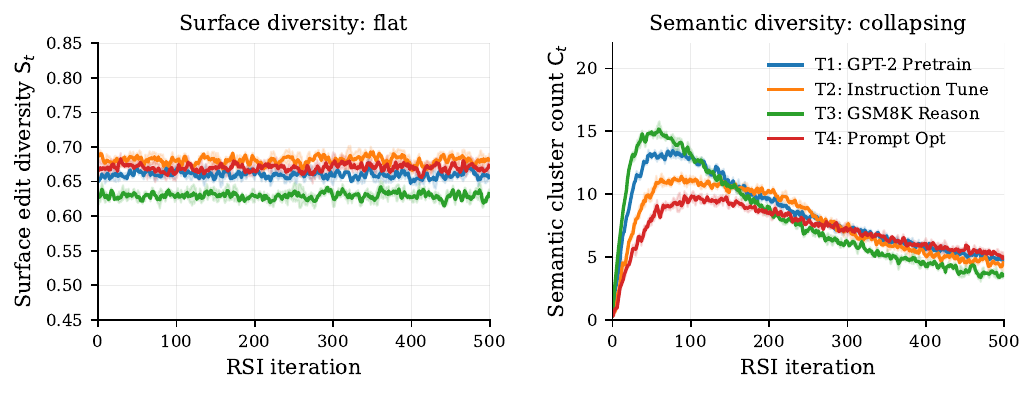}
  \vspace{-3mm}
  \caption{\textbf{Algorithmic mode collapse in code-level ARLs.} Left: surface-level edit diversity, measured as mean pairwise normalized edit distance over a sliding window of 20 accepted diffs, remains essentially flat across $500$ iterations on all four tasks. Right: semantic-cluster count, obtained by Sentence-BERT \citep{reimers2019sbert} embeddings of LLM-summarized edit descriptions followed by HDBSCAN clustering \citep{campello2013hdbscan}, collapses by 50 to 70\% over the same horizon. The agent keeps editing different lines of code, but is increasingly proposing the same \emph{kind} of change. Shaded regions are $\pm 1$ standard deviation over 3 seeds.}
  \label{fig:money}
\end{figure*}
 
However, we find that this expectation should be questioned. Code-level ARL, despite using a verifiable external metric, exhibits a previously undocumented failure mode we call \textbf{algorithmic mode collapse}: the \emph{surface} of the agent's behavior, such as the lines it edits, the lexical form of its diffs, remains diverse, while the \emph{semantics} of its proposals, what each edit is actually trying to accomplish, progressively concentrates on a small number of recurring patterns. Figure~\ref{fig:money} shows the phenomenon at a glance: across four NLP-relevant tasks, surface edit diversity is essentially flat over $500$ iterations, while the number of semantic clusters in the proposal stream collapses by $50$ to $70\%$. This narrowing changes the character of the loop. Rather than continuing to search broadly over possible algorithmic interventions, the proposer increasingly returns to a small repertoire of edits that reliably improve the in-loop metric. Such concentration might be benign if it reflected convergence on genuinely useful mechanisms. In our experiments, however, it is accompanied by a growing gap between optimization-metric improvements claimed inside the loop and improvements measured on independent held-out evaluations. As the proposer's repertoire narrows, it increasingly produces edits that overfit the in-loop signal rather than improve the underlying system. By iteration $300$, average in-loop gains overstate held-out gains by a factor of $2.0$ to $2.6$ across our tasks.
 
Building on this diagnosis, we introduce the mitigation strategy Diversity-Aware Proposal Sampling (\textsc{DAPS}), that injects three lightweight components into an existing ARL loop: a category-coverage reweighting that nudges the proposer toward under-represented edit mechanisms, a persistent semantic memory that suppresses proposals too similar to recently accepted ones, and a validation gate that reverts the pipeline to its last \emph{faithful} state when the gap between the in-loop metric and a sparingly consulted audit metric exceeds a calibrated threshold. \textsc{DAPS} reduces semantic-cluster decay of edit descriptions by $69.1\%$ and improves the relative faithfulness by $81.6\%$ averaged across tasks, while matching or slightly trailing vanilla autoresearch on in-loop optimization speed. Because the gate consumes the audit metric, we also score every configuration on blind sets that no component ever accesses, where the advantage persists.
 
Our contributions can be summarized as follows:
(i) We conduct the first controlled empirical study of code-level ARL dynamics over long horizons across NLP tasks, producing ARL trajectories with logged proposals, diffs, executions, and held-out evaluations. 
(ii) We introduce a four-axis diagnostic instrument: surface, semantic, mechanism, and corpus-similarity diversity, together with a faithfulness audit that operationalize \emph{algorithmic mode collapse} and reveal its link to in-loop/held-out divergence.
(iii) We propose \textsc{DAPS}, a drop-in mitigation that materially reduces collapse and improves faithfulness across three proposer LLMs without slowing optimization.
(iv) We show that surface-only diversity measurements that are common in prior data-level collapse analyses \citep{rdiverse2026,prism2026}, systematically miss algorithmic mode collapse, motivating semantic-aware monitoring for future ARL deployment.
 
\section{Related Work}
\label{sec:related}
\textbf{Autonomous Research Agents.}
The AI-Scientist \citep{lu2024aiscientist}, Agent Laboratory \citep{schmidgall2025agentlab}, AI-Researcher \citep{airesearcher2025}, and AI Co-Scientist \citep{gottweis2025coscientist} systems orchestrate LLM agents through hypothesis generation, experimentation, and writeup. \citet{karpathy2026autoresearch} distilled the core experiment-loop pattern into a 630-line script that operates at single-GPU scale, popularizing the term \emph{autoresearch} and motivating a wave of variants \citep{walker2026autoresearch_tools}. Concurrently, \citet{whyllmsnotsci2026} document six recurring failure modes across four end-to-end autonomous ML research attempts, while \citet{autoresearchbench2026} report that even frontier models reach only $9.4\%$ accuracy on rigorous scientific-literature discovery. Our work is complementary: where prior efforts evaluate end-to-end research quality, we analyze the \emph{dynamics} of the experimentation loop.
 
\textbf{Diversity Collapse in Recursive Training.}
\citet{shumailov2024curse} established that recursively training on model-generated data leads to distributional narrowing. Subsequent work has refined this picture for self-play and self-training \citep{recursiveloops2025,dces2026}. Most directly related, \citet{rdiverse2026} identify a \emph{Diversity Illusion} in Challenger and Solver self-play. i.e., surface variation persists while underlying patterns collapse. \citet{prism2026} diagnose \emph{curriculum collapse} in self-evolving reasoning systems, proposing semantic cluster-coverage rewards. We adopt their semantic-vs-surface framing but transport it to a fundamentally different setting: the optimization signal in code-level ARL is an external execution, not a learned judge or self-generated label, so prior collapse mechanisms (model-on-its-own-data) do not directly apply. Our diagnosis identifies a distinct mechanism rooted in the \emph{proposer's} prior distribution interacting with metric overfitting.
 
\textbf{Reward Hacking and Goodharting.}
That optimizers exploit proxy metrics is a classical observation \citep{christiano2017deeprl,skalse2022reward,paneffects}. Closer to our setting, \citet{gao2023scaling} document that the gap between proxy and gold rewards widens with optimization pressure. Algorithmic mode collapse can be read as a Goodhart-style phenomenon at the \emph{proposal} layer: a narrowing repertoire of edits is selected precisely because it reliably moves the in-loop metric, including when the underlying system has not improved.

 
\textbf{Plagiarism of AI-Generated Research.}
\citet{gupta2025glitters} show that AI-generated research ideas frequently recapitulate prior work without attribution. We operationalize a related diagnosis, asking whether a proposed modification is essentially retrieved from the proposer's training data.
 
\section{Methodology}
\label{sec:method}
 
\subsection{Code-Level ARL: Setup and Notation}
\label{sec:setup}
 
A code-level ARL loop is a tuple $(\mathcal{P}, \pi_\theta, \mathcal{E}, R)$ where $\mathcal{P}$ is a training pipeline represented as a set of source files, $\pi_\theta$ is an LLM proposer, $\mathcal{E}$ is an executor that runs a candidate pipeline to obtain a metric value, and $R: \mathbb{R} \to \{0,1\}$ is an acceptance rule. At iteration $t$, the proposer samples a textual modification description $d_t \sim \pi_\theta(\cdot \mid \mathcal{P}_t, h_t)$ together with a code patch $\rho_t$, where $h_t$ summarizes the loop history. The executor evaluates $\mathcal{P}_t \oplus \rho_t$ on an optimization metric $m^{\text{opt}}$; the new pipeline is retained if $R(m^{\text{opt}}(\mathcal{P}_t \oplus \rho_t) - m^{\text{opt}}(\mathcal{P}_t)) = 1$, typically when the metric improves by more than a noise threshold $\epsilon$. Crucially, $m^{\text{opt}}$ is what the agent sees; \S\ref{sec:faith} defines the two evaluations that it does \emph{not} optimize.
 
\subsection{Multi-Axis Diversity Instrumentation}
\label{sec:axes}
Existing analyses of diversity collapse measure it primarily through token- or embedding-level signals over generated \emph{text} \citep{rdiverse2026,prism2026}. In a code-level autonomous research loop, the natural unit is the edit, which admits multiple, partially independent notions of diversity. We instrument four axes as follows.
 
\textbf{Surface diversity ($\mathrm{S}_t$).} For each accepted edit, we extract its unified diff and compute the normalized Levenshtein distance between every pair of diffs in a sliding window of the $W=20$ most recent accepted edits. $\mathrm{S}_t$ is the window-mean. This captures whether the agent is editing different lines of code in lexically different ways.
 
\textbf{Semantic diversity ($\mathrm{C}_t$).} The proposer emits, alongside each patch, a one-sentence natural-language description of its intent (or we elicit one post-hoc from a fixed summarizer LLM). We embed all descriptions accumulated up to iteration $t$ with Sentence-BERT \citep{reimers2019sbert} and cluster them with HDBSCAN \citep{campello2013hdbscan} at a fixed minimum cluster size of $5$. $\mathrm{C}_t$ is the number of clusters whose youngest member was added within the last $W$ iterations. The motivation is direct: \emph{what} the agent is trying to accomplish, like learning-rate adjustment, attention re-formulation, regularization tweak, is exactly what surface metrics fail to capture.
 
\textbf{Mechanism entropy ($\mathrm{H}_t$).} Surface and semantic axes are continuous; we add a discrete taxonomy for interpretability. To derive the taxonomy we conducted a pilot study on $187$ edit descriptions sampled uniformly from publicly available code-level ARL logs, embedded with Sentence-BERT and clustered with affinity propagation \citep{frey2007affinity}; resulting clusters were iteratively merged when their natural-language summaries overlapped, yielding nine categories that jointly cover $96.4\%$ of pilot edits (a residual ``other'' bucket absorbs the long tail). Because the pilot logs come from other ARL systems, the categories are frozen before any of our own runs. Table~\ref{tab:taxonomy} lists the categories with a representative paraphrased example each. At loop time each edit description is assigned by a rule-based parser over a curated keyword inventory, cross-validated against parallel labelling by a frontier LLM annotator; Cohen's $\kappa$ between the two on a $200$-edit held-out validation set is $0.84$, indicating almost perfect agreement \citep{landis1977kappa}. $\mathrm{H}_t$ is the Shannon entropy~\citep{shannon1948mathematical} of the category distribution over a window of $W$ accepted edits. Low $\mathrm{H}_t$ in the presence of high $\mathrm{S}_t$ is the signature of algorithmic mode collapse: the agent edits diverse lines of code but is increasingly trying to do the same kind of thing. Appendix~\ref{app:taxonomy} shows the collapse to be invariant to taxonomy granularity and to human labelling.
 
\begin{table}[t]
\centering
\small
\resizebox{0.48\textwidth}{!}{
\begin{tabular}{ll}
\toprule
Category & Representative edit (paraphrased) \\
\midrule
Optimizer        & Switch AdamW to Lion; adjust $\beta_2$  \\
Scheduling       & Lengthen LR warmup; set decay floor      \\
Architecture     & Replace LayerNorm with RMSNorm           \\
Data             & Filter by sequence length; rebalance mix \\
Loss             & Add label-smoothing or auxiliary term    \\
Regularization   & Raise attention-projection dropout       \\
Numerical        & FP32 softmax; revise grad-clip bound     \\
Decoding         & Adjust top-$p$ / nucleus threshold       \\
Other            & Logging, telemetry, build-script edits   \\
\bottomrule
\end{tabular}}
\caption{\textbf{Mechanism taxonomy} used for $\mathrm{H}_t$. Categories were derived from a pilot study (see text) and cover $96.4\%$ of observed edits.}
\label{tab:taxonomy}
\end{table}
 
\textbf{Corpus similarity ($\mathrm{V}_t$).} Each edit description is encoded and matched, via nearest-neighbor cosine similarity, against a corpus of $1.1$M arXiv cs.CL/cs.LG abstracts published before January 2024. $\mathrm{V}_t$ is the mean top-1 similarity over the window. We treat this axis as descriptive: a rising $\mathrm{V}_t$ is consistent with the proposer leaning on familiar, frequently described patterns rather than exploring \citep{gupta2025glitters}, but it does not by itself establish retrieval, since the same mechanism can be phrased in standard or in unusual ML vocabulary and the corpus is abstracts only, pre-2024, and field-restricted. 
 
\subsection{The Three-Tier Faithfulness Audit}
\label{sec:faith}
We separate three evaluation roles. The \emph{in-loop} metric $m^{\text{opt}}$ is the only signal the proposer optimizes and the acceptance rule consumes. The \emph{audit} metric $m^{\text{audit}}$, defined on data the agent cannot infer from $\mathcal{P}_t$ and targeting the \emph{intended} capability rather than the proxy, is never observed by the proposer and is read only by the gate of \S\ref{sec:daps} and its threshold calibration, every $K=10$ iterations, through a scalar test whose only effects are a revert and a templated, value-free notice ($2.7 \pm 0.9$ reverts per run). Since it can thus influence the trajectory, we call it a validation signal rather than a fully held-out test. The \emph{blind} metric $m^{\text{blind}}$ is read by no component at any point, gate and calibration included, and is evaluated once per run; it is our headline faithfulness evaluation (partitions in Appendix~\ref{app:tiers}).

Every $K=10$ iterations, we evaluate the best-so-far pipeline on $m^{\text{audit}}$ and compare against $m^{\text{opt}}$. Concretely, we define the \emph{faithfulness gap}
\begin{equation}
  \Delta^{\text{faith}}_t = \underbrace{\bar{g}^{\text{opt}}_t}_{\text{in-loop gain}} - \underbrace{\bar{g}^{\text{audit}}_t}_{\text{audited gain}},
  \label{eq:faith}
\end{equation}
where each gain is the improvement over the starting pipeline, signed so that positive means better. A widening $\Delta^{\text{faith}}_t$ implies that the loop is increasingly optimizing the proxy without commensurate downstream effect; $\Delta^{\text{blind}}_T$ is defined analogously.
 
\subsection{\textsc{DAPS}: Diversity-Aware Proposal Sampling}
\label{sec:daps}
To mitigate such identified algorithmic mode collapse, we propose the 
\textsc{DAPS} framework which composes three components that target the diagnosed pathology directly.
 
\textbf{Category-Coverage Reweighting (\textsc{ccr}).} Let $p_t(c)$ denote the empirical frequency of mechanism category $c$ in the last $W$ accepted edits. When sampling proposals we draw $N$ candidates from $\pi_\theta$, classify them on the fly, and re-weight them by $w(\rho) = \exp\big(-\log p_t(c(\rho)) / \tau_c\big)$ before acceptance. \textsc{ccr} does not alter the executor; it only changes which candidate is presented for execution. The motivation is that the proposer's prior over edit \emph{types} is sharply peaked toward optimizer/learning-rate edits, and acceptance pressure amplifies skew.
 
\textbf{Persistent Edit Memory (\textsc{pem}).} We maintain a first-in-first-out (FIFO) buffer of the Sentence-BERT embeddings of the last $M=200$ accepted edit descriptions and reject any candidate whose cosine similarity to its nearest neighbor in memory exceeds $\tau_m$. \textsc{pem} prevents the loop from cycling on near-duplicate semantics regardless of whether the underlying code differs. This is the code-level analog of the memory-augmented penalty of \citet{rdiverse2026}, applied to edit descriptions rather than self-play questions.
 
\textbf{Audit-Based Validation Gate (\textsc{hvg}).} Every $K=10$ iterations we compute $\Delta^{\text{faith}}_t$ from Eq.~\ref{eq:faith}. If $\Delta^{\text{faith}}_t$ exceeds a task-calibrated threshold $\tau_h$, the gate reverts $\mathcal{P}$ to the last checkpoint with $\Delta^{\text{faith}} \le \tau_h$ and feeds a structured failure summary back to the proposer. \textsc{hvg} is the only component that consumes the audit metric; it does so sparingly, through the scalar test of \S\ref{sec:faith}, and is the mechanism by which Goodhart-style edits are eventually undone.
 
The full procedure has been summarized in Algorithm~\ref{alg:daps}. 
 
\begin{algorithm}[t]
\caption{Diversity-Aware Proposal Sampling}
\label{alg:daps}
\begin{algorithmic}[1]
\State \textbf{Input:} pipeline $\mathcal{P}_0$, proposer $\pi_\theta$, executor $\mathcal{E}$, metrics $m^{\text{opt}}, m^{\text{audit}}$
\State Initialize memory $\mathcal{M} \gets \emptyset$, history $h \gets \emptyset$
\For{$t = 1, \ldots, T$}
  \State Sample $N$ candidates $\{(\rho_i, d_i)\} \sim \pi_\theta(\cdot \mid \mathcal{P}_{t-1}, h)$
  \State Compute $w_i \gets \exp(-\log p_t(c(\rho_i))/\tau_c)$ \Comment{\textsc{ccr}}
  \State Drop $\rho_i$ if $\max_{e\in\mathcal{M}} \cos(\phi(d_i), e) > \tau_m$ \Comment{\textsc{pem}}
  \State Choose $\rho^\star \gets$ weighted-sample remaining candidates by $w_i$
  \State $m \gets \mathcal{E}(\mathcal{P}_{t-1} \oplus \rho^\star)$
  \If{$R(m - m^{\text{opt}}(\mathcal{P}_{t-1})) = 1$}
    \State $\mathcal{P}_t \gets \mathcal{P}_{t-1} \oplus \rho^\star$; add $\phi(d^\star)$ to $\mathcal{M}$
  \Else
    \State $\mathcal{P}_t \gets \mathcal{P}_{t-1}$
  \EndIf
  \If{$t \bmod K = 0$ \textbf{and} $\Delta^{\text{faith}}_t > \tau_h$} \Comment{\textsc{hvg}}
    \State Revert $\mathcal{P}_t$ to last faithful checkpoint
    \State Append failure summary to $h$
  \EndIf
\EndFor
\end{algorithmic}
\end{algorithm}
 
\section{Experiments}
\label{sec:exp}
\subsection{Experiment Setup} 
\subsubsection{Tasks and Datasets}
\label{sec:tasks}
We instantiate code-level ARL on four NLP-relevant tasks spanning pretraining, post-training, reasoning, and inference-time configuration.
 
\textbf{T1: Small-LM Pretraining.} A GPT-2-small style model \citep{radford2019gpt2} ($\sim$$124$M parameters) trained on a $0.5$B-token subset of OpenWebText for a fixed $5$-minute budget per execution, following the \texttt{autoresearch} setup of \citet{karpathy2026autoresearch}. $m^{\text{opt}}$: in-distribution validation loss. $m^{\text{audit}}$: perplexity on LAMBADA \citep{paperno2016lambada} and a held-out C4 \citep{raffel2020t5} subset.
 
\textbf{T2: Instruction Tuning.} A Llama-3.2-1B base model \citep{dubey2024llama3} fine-tuned on a $20$k-example Alpaca subset \citep{taori2023alpaca} via LoRA. $m^{\text{opt}}$: held-in instruction-following win-rate (AlpacaEval-style \citep{li2023alpacaeval} against a fixed reference) on a $500$-example development split. $m^{\text{audit}}$: MMLU \citep{hendrycks2021mmlu} 5-shot accuracy and the IFEval prompt-following benchmark \citep{zhou2023ifeval}.
 
\textbf{T3: Reasoning Fine-Tuning.} A Qwen-2.5-1.5B \citep{qwen25} model fine-tuned on GSM8K \citep{cobbe2021gsm8k} training split. $m^{\text{opt}}$: GSM8K dev-set exact-match accuracy. $m^{\text{audit}}$: ARC-Easy \citep{clark2018arc} and MATH-500 \citep{hendrycks2021math} subset accuracy.
 
\textbf{T4: Prompt Optimization.} Frozen Llama-3.2-3B prompted on a question-answering subset, where the autonomous research loop edits a Python DSPy-style prompt program \citep{khattab2023dspy}. $m^{\text{opt}}$: dev-set accuracy on a sampled ARC-Easy split. $m^{\text{audit}}$: ARC-Challenge \citep{clark2018arc} and CommonsenseQA \citep{talmor2019commonsenseqa}.
 
The $m^{\text{opt}}$ and $m^{\text{audit}}$ are computed with independent prompts and data partitions. Here, audit tasks are chosen to be  \emph{capability-overlapping but distribution-different} relative to in-loop ones: a genuine capability improvement should transfer, while a metric-specific overfitting should not. The blind sets, evaluated once per run and read by no component of the loop, are WikiText-103 log-perplexity \citep{merity2017wikitext} for T1, win-rate on $500$ held-out Dolly instructions \citep{conover2023dolly} under the same judging protocol for T2, SVAMP accuracy \citep{patel2021svamp} for T3, and OpenBookQA accuracy \citep{mihaylov2018openbookqa} for T4. 
 
\subsubsection{Baselines}
\label{sec:baselines}
We compare \textsc{DAPS} against eight baselines spanning the major existing strategies.
(\textbf{B1}) \textsc{Vanilla AR}: a faithful reimplementation of \citet{karpathy2026autoresearch} with the same proposer LLM. (\textbf{B2}) \textsc{HiTemp}: vanilla autoresearch with proposer temperature raised from $0.7$ to $1.2$, a frequently suggested ad-hoc fix for low diversity. (\textbf{B3}) \textsc{R-Diverse-A}: vanilla autoresearch augmented with the Memory-Augmented Penalty of \citet{rdiverse2026}, applied to edit descriptions (the strongest published mitigation transposed to our setting). (\textbf{B4}) \textsc{Prism-A}: vanilla autoresearch with the semantic cluster-coverage reward of \citet{prism2026}. (\textbf{B5}) \textsc{Reflexion}: vanilla autoresearch with an additional reflective summary \citep{shinn2023reflexion} prepended to $h$ at every step. (\textbf{B6}) \textsc{RandSearch}: a proposer-free baseline that samples edits uniformly from a corpus of code modifications harvested from the proposer LLM in iteration $1$; this isolates the contribution of LLM's adaptive proposals beyond a static edit distribution. (\textbf{B7}) \textsc{HO-EarlyStop}: vanilla autoresearch that reads $m^{\text{audit}}$ every $K{=}10$ iterations and returns the best-audit checkpoint. (\textbf{B8}) \textsc{HO-Revert}: audit-based reversion alone, that is \textsc{hvg} without \textsc{ccr} or \textsc{pem}. B7 and B8 match \textsc{DAPS} in audit frequency, compute, and feedback format, while \textsc{HiTemp} and \textsc{RandSearch} are diagnostic controls rather than competitors. All baselines share \textsc{DAPS}'s tuning budget (Appendix~\ref{app:grids}).
 
\subsubsection{Evaluation Metrics}
\label{sec:metrics}
For each method/task/seed trajectory, we report:
(i) \textit{In-loop gain} $\bar{g}^{\text{opt}}_T$;
(ii) \textit{Audited gain} $\bar{g}^{\text{audit}}_T$ and \textit{blind gain} $\bar{g}^{\text{blind}}_T$;
(iii) \textit{Faithfulness ratios} $\rho^{\text{audit}}_T = \bar{g}^{\text{audit}}_T / \bar{g}^{\text{opt}}_T$ and $\rho^{\text{blind}}_T = \bar{g}^{\text{blind}}_T / \bar{g}^{\text{opt}}_T$ (closer to $1$ is better; $<1$ indicates Goodharting), where all gains are signed improvements oriented so that positive means better: on T1 numerator and denominator are both reductions in nats per token ($0.058/0.142 = 0.41$ under \textsc{Vanilla AR}), on T2 to T4 both are percentage points, and ratios are reported only when $\bar{g}^{\text{opt}}_T > \epsilon$, as held throughout. We write $\rho_T$ without a superscript when a statement holds for both;
(iv) \textit{Faithfulness gap} $\Delta^{\text{faith}}_T$ at end of run (Eq.~\ref{eq:faith});
(v) The four diagnostic axes of \S\ref{sec:axes} evaluated at end of run: \textit{surface diversity} $\mathrm{S}_T$, \textit{semantic cluster count} $\mathrm{C}_T$, \textit{mechanism entropy} $\mathrm{H}_T$, and \textit{corpus similarity} $\mathrm{V}_T$;
(vi) \textit{Cluster decay} $\Delta\mathrm{C} = (\mathrm{C}_{T/5} - \mathrm{C}_T)/\mathrm{C}_{T/5}$, the fractional drop in clusters from the early-warm-up to the final window, used as a scalar summary of trajectory-level collapse.
We report mean $\pm$ s.d. over $3$ seeds for the main configurations and aggregate $5$ seeds for T1 (where per-trajectory variance is largest).

\begin{table*}[t]
\centering
\small
\setlength{\tabcolsep}{4pt}
\begin{tabular}{lcccccccc}
\toprule
& \multicolumn{2}{c}{\textbf{T1 (Pretrain)}} & \multicolumn{2}{c}{\textbf{T2 (InstrTune)}} & \multicolumn{2}{c}{\textbf{T3 (Reason)}} & \multicolumn{2}{c}{\textbf{T4 (Prompt)}}\\
\cmidrule(lr){2-3}\cmidrule(lr){4-5}\cmidrule(lr){6-7}\cmidrule(lr){8-9}
Method & $\bar{g}^{\text{opt}}$ & $\rho^{\text{audit}}_T$ & $\bar{g}^{\text{opt}}$ & $\rho^{\text{audit}}_T$ & $\bar{g}^{\text{opt}}$ & $\rho^{\text{audit}}_T$ & $\bar{g}^{\text{opt}}$ & $\rho^{\text{audit}}_T$\\
\midrule
\textsc{Vanilla AR}        & $+0.142_{\pm.011}$ & $0.41_{\pm.05}$ & $+8.9_{\pm1.2}$ & $0.46_{\pm.07}$ & $+11.7_{\pm1.8}$ & $0.38_{\pm.06}$ & $+9.4_{\pm1.1}$ & $0.49_{\pm.05}$ \\
\textsc{HiTemp}            & $+0.131_{\pm.014}$ & $0.43_{\pm.06}$ & $+8.4_{\pm1.4}$ & $0.49_{\pm.06}$ & $+10.9_{\pm2.1}$ & $0.40_{\pm.08}$ & $+9.0_{\pm1.3}$ & $0.51_{\pm.06}$ \\
\textsc{R-Diverse-A}       & $+0.128_{\pm.012}$ & $0.55_{\pm.05}$ & $+8.1_{\pm1.0}$ & $0.58_{\pm.06}$ & $+10.2_{\pm1.6}$ & $0.51_{\pm.07}$ & $+8.7_{\pm1.2}$ & $0.60_{\pm.05}$ \\
\textsc{Prism-A}           & $+0.135_{\pm.010}$ & $0.59_{\pm.04}$ & $+8.6_{\pm1.1}$ & $0.61_{\pm.05}$ & $+10.8_{\pm1.5}$ & $0.54_{\pm.06}$ & $+9.1_{\pm1.0}$ & $0.62_{\pm.04}$ \\
\textsc{Reflexion}         & $+0.138_{\pm.013}$ & $0.47_{\pm.06}$ & $+8.7_{\pm1.3}$ & $0.50_{\pm.07}$ & $+11.2_{\pm1.7}$ & $0.43_{\pm.07}$ & $+9.3_{\pm1.1}$ & $0.53_{\pm.05}$ \\
\textsc{RandSearch}        & $+0.073_{\pm.018}$ & $0.78_{\pm.09}$ & $+4.1_{\pm1.5}$ & $0.81_{\pm.10}$ & $+5.6_{\pm1.9}$ & $0.74_{\pm.11}$ & $+5.0_{\pm1.4}$ & $0.79_{\pm.09}$ \\
\midrule
\rowcolor{tableaublue!20} \textsc{DAPS} (ours)       & $\mathbf{+0.139_{\pm.009}}$ & $\mathbf{0.78_{\pm.04}}$ & $\mathbf{+8.8_{\pm1.0}}$ & $\mathbf{0.80_{\pm.05}}$ & $\mathbf{+11.5_{\pm1.4}}$ & $\mathbf{0.74_{\pm.06}}$ & $\mathbf{+9.3_{\pm0.9}}$ & $\mathbf{0.82_{\pm.04}}$ \\
\bottomrule
\end{tabular}
\caption{\textbf{Main results.} In-loop gain $\bar{g}^{\text{opt}}$ and audited faithfulness ratio $\rho^{\text{audit}}_T$ at $T=300$ across four tasks. Both are oriented so that higher is better: $\bar{g}^{\text{opt}}$ is the loss reduction in nats per token on T1 and the win-rate or accuracy gain in percentage points on T2 to T4. $\rho^{\text{audit}}_T$ closer to $1$ indicates that in-loop gains transfer to the audit evaluation; blind-set counterparts are in Table~\ref{tab:tiers} and Appendix~\ref{app:tiers}. Subscripts denote $\pm 1$ standard deviation over $3$ seeds (5 for T1).}
\vspace{-3mm}
\label{tab:main}
\end{table*}

\subsubsection{Implementation Details}
\label{sec:impl}
In experiments, our primary pipeline forks the public \texttt{autoresearch} repository \citep{karpathy2026autoresearch} and adds task adapters, the four diagnostic axes, and the \textsc{DAPS} components. To verify that our findings are not artifacts of a single implementation, we additionally instantiate the same ARL specification on top of Aider \citep{aider2024gauthier}, a popular open-source code-editing agent whose proposal mechanism differs substantially from \texttt{autoresearch}: Aider emits structured \texttt{SEARCH}/\texttt{REPLACE} blocks committed via git rather than direct unified diffs, manages its own multi-file repository map and chat history, and decouples ``decide what to change'' from ``apply the change'' through separate LLM passes. We wrap Aider as a drop-in proposer-and-editor module within our loop while keeping the executor, acceptance rule, history summary, and diagnostic instrumentation identical. The diagnostic axes are computed in exactly the same way on both frameworks: edit descriptions are extracted from Aider's commit messages (which it auto-generates) and re-summarized by the same fixed summarizer LLM for consistency with the \texttt{autoresearch} pipeline. The proposer LLM for our main configuration is Claude Opus 4.7 with $\text{temperature}{=}0.7$, sampled through the Anthropic API; robustness ablations additionally use GPT-5.2 and Llama-3.3-70B served via vLLM. For semantic embedding we use \texttt{all-mpnet-base-v2} \citep{reimers2019sbert}; clustering uses HDBSCAN with $\text{min\_cluster\_size}=5$ and $\text{min\_samples}=2$ \citep{campello2013hdbscan}; UMAP \citep{mcinnes2018umap} is used only for visualization. Hyperparameters of \textsc{DAPS} are fixed across all tasks and both frameworks at $\tau_c{=}1.0$, $\tau_m{=}0.85$, $M{=}200$, $K{=}10$, and a task-calibrated $\tau_h$ chosen on the first $30$ iterations to be the $80$th percentile of $|\Delta^{\text{faith}}|$ observed under \textsc{Vanilla AR}. 
Each ARL execution is bounded at $5$ minutes on a single A100; full sweeps used approximately $6{,}200$ A100-hours. 
 
\subsection{Main Results and Analysis}
\label{sec:res_main}
We report our main results in Table~\ref{tab:main}. Three findings stand out.
\textit{First, vanilla autoresearch overfits.} Across all four tasks, $\rho^{\text{audit}}_T$ for \textsc{Vanilla AR} ranges from $0.38$ to $0.49$, meaning roughly half to two-thirds of the gain claimed inside the loop fails to materialize on held-out evaluation. This is the quantitative form of the gap previewed in \S\ref{sec:intro}.
\textit{Second, data-level diversity mitigations transfer only partially.} \textsc{R-Diverse-A} and \textsc{Prism-A}, the strongest published collapse mitigations from the data-level literature, improve faithfulness ($\rho^{\text{audit}}_T$ rises from $\approx 0.43$ to $\approx 0.58$ averaged across tasks) but neither fully closes the gap nor preserves in-loop performance ($\bar{g}^{\text{opt}}$ degrades by $3\%$ to $13\%$ relative to \textsc{Vanilla AR} on T2 and T3). This supports our claim that code-level ARL involves a distinct failure mode requiring code-level instrumentation.
\textit{Third, raising temperature is not a remedy.} \textsc{HiTemp} achieves marginally higher faithfulness at marginally lower in-loop gain, but the effect is well within the standard deviation. Increased proposer entropy does not, in our experiments, translate to increased \emph{semantic} diversity, a phenomenon we examine in \S\ref{sec:res_collapse}.
\textsc{DAPS} achieves the highest $\rho^{\text{audit}}_T$ on all four tasks while remaining within one standard deviation of the best $\bar{g}^{\text{opt}}$. We attribute the absence of an in-loop cost to two facts: \textsc{ccr}/\textsc{pem} reject duplicate proposals \emph{before} executor calls, so they do not waste budget; and \textsc{hvg} reverts only when in-loop and audit metrics diverge, so a faithful trajectory is unaffected.

\begin{table}[t]
\centering
\small
\setlength{\tabcolsep}{4pt}
\begin{tabular}{lccccc}
\toprule
Method & audit & $\bar{g}^{\text{opt}}$ & $\rho^{\text{aud}}_T$ & $\rho^{\text{bli}}_T$ & $\Delta\mathrm{C}$ \\
\midrule
\textsc{Vanilla AR}          & none     & $1.000$ & $0.44$ & $0.41$ & $0.68$ \\
\textsc{R-Diverse-A}         & none     & $0.900$ & $0.56$ & $0.52$ & $0.39$ \\
\textsc{Prism-A}             & none     & $0.950$ & $0.59$ & $0.55$ & $0.34$ \\
\textsc{ccr}+\textsc{pem}    & none     & $0.986$ & $0.71$ & $0.67$ & $0.28$ \\
\textsc{HO-EarlyStop}        & $K{=}10$ & $0.910$ & $0.62$ & $0.57$ & $0.66$ \\
\textsc{HO-Revert}           & $K{=}10$ & $0.953$ & $0.67$ & $0.62$ & $0.55$ \\
\rowcolor{tableaublue!20}\textsc{DAPS} (full) & $K{=}10$ & $0.983$ & $\mathbf{0.79}$ & $\mathbf{0.74}$ & $\mathbf{0.21}$ \\
\bottomrule
\end{tabular}
\caption{Faithfulness under matched audit access at $T=300$, averaged across tasks and seeds, with $\bar{g}^{\text{opt}}$ normalized by \textsc{Vanilla AR} and $\rho^{\text{aud}}$, $\rho^{\text{bli}}$ abbreviating $\rho^{\text{audit}}$, $\rho^{\text{blind}}$. Methods in the upper block never read any external metric; those in the lower block read $m^{\text{audit}}$ every $K$ iterations under the same evaluation frequency, compute, and feedback format.}
\vspace{-3mm}
\label{tab:tiers}
\end{table}

\subsection{Faithfulness Under Matched Audit Access}
\label{sec:res_tiers}
Table~\ref{tab:tiers} studies whether the faithfulness advantage survives on data the gate never touches, and whether it follows from the method or from audit access that no baseline is granted. Every configuration loses only $0.03$ to $0.05$ between the audit and the blind tier, including the four that read no external metric at all, so that offset reflects benchmark idiosyncrasy rather than leakage through the gate: \textsc{DAPS} keeps its advantage blind ($0.74$ against $0.79$ audited), and per task the relative improvement over \textsc{Vanilla AR} is $83.7\%$ blind against $81.6\%$ audited. On equal footing, three conclusions follow. \emph{First}, with zero audit access \textsc{ccr}+\textsc{pem} already surpasses both published diversity mitigations on both faithfulness tiers and on $\Delta\mathrm{C}$ at a smaller in-loop cost. \emph{Second}, audit access alone reaches at most $\rho^{\text{audit}}_T = 0.67$ ($0.62$ blind) and barely reduces collapse ($\Delta\mathrm{C} \ge 0.55$), so the extra signal explains only part of the advantage. \emph{Third}, the two ingredients are complementary, as \S\ref{sec:daps} intends. Appendix~\ref{app:validity} indexes the validity, robustness, and scope checks behind these numbers.
\vspace{-1mm}
 
\begin{figure*}[t]
  \centering
  \includegraphics[width=0.9\linewidth]{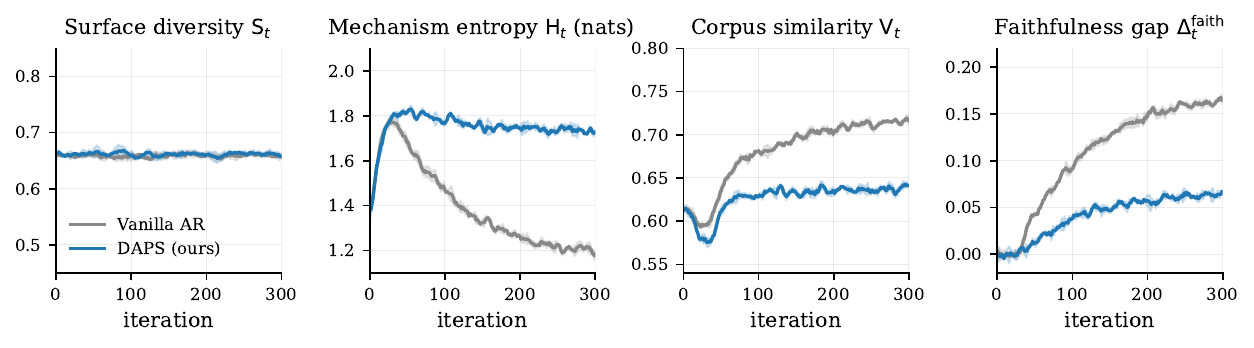}
  \vspace{-5mm}
  \caption{Diagnostic axes over iterations on T2. Mechanism entropy $\mathrm{H}_t$ collapses by $\sim$$0.6$ nats while surface diversity $\mathrm{S}_t$ remains flat. Corpus similarity $\mathrm{V}_t$ rises, which is consistent with reliance on patterns close to pretraining text; the in-loop/audit gap $\Delta^{\text{faith}}_t$ widens monotonically under \textsc{Vanilla AR} and is materially shrunk under \textsc{DAPS}.}
  \vspace{-3mm}
  \label{fig:diagnostic}
\end{figure*}
 
\subsection{Dissecting the Collapse}
\label{sec:res_collapse}
Figure~\ref{fig:diagnostic} traces our four diagnostic axes on T2 under \textsc{Vanilla AR} and \textsc{DAPS}. We have three observations as follows.
(a) \emph{Surface diversity is a poor warning sign.} $\mathrm{S}_t$ stays in $[0.66, 0.68]$ for both methods throughout, while $\mathrm{H}_t$ under \textsc{Vanilla AR} drops from $1.76$ nats at $t=30$ to $1.18$ nats at $t=300$. Any monitor that watched only $\mathrm{S}_t$ would have raised no flag.
(b) \emph{Corpus similarity rises with collapse.} $\mathrm{V}_t$ increases from $0.60$ to $0.72$ under \textsc{Vanilla AR}, so the agent concentrates on edits whose descriptions are increasingly close to commonly published ML text \citep{gupta2025glitters}. This is consistent with the proposer relying on its prior rather than exploring, although as a descriptive statistic it does not on its own establish retrieval (\S\ref{sec:axes}).
(c) \emph{Faithfulness erodes monotonically.} $\Delta^{\text{faith}}_t$ grows from $\approx 0.05$ at $t=60$ to $\approx 0.17$ at $t=300$ under \textsc{Vanilla AR}, mirroring the entropy collapse with almost no iteration lag. Under \textsc{DAPS}, the gap stabilizes at $\approx 0.07$.
Table~\ref{tab:diagaxes} reports the axis statistics across all tasks and methods. The pattern is consistent: methods that close the entropy gap (\textsc{Prism-A}, \textsc{DAPS}) also close the faithfulness gap; methods that do not (\textsc{HiTemp}, \textsc{Reflexion}) do not. Surface diversity correlates with neither.
 
\begin{table}[t]
\centering
\small
\begin{tabular}{lcccc}
\toprule
Method & $\mathrm{S}_T$ & $\mathrm{H}_T$ & $\Delta\mathrm{C}$ & $\Delta^{\text{faith}}_T$ \\
\midrule
\textsc{Vanilla AR}   & $0.66$ & $1.18$ & $0.68$ & $0.171$ \\
\textsc{HiTemp}       & $0.69$ & $1.24$ & $0.61$ & $0.158$ \\
\textsc{R-Diverse-A}  & $0.67$ & $1.51$ & $0.39$ & $0.108$ \\
\textsc{Prism-A}      & $0.65$ & $1.59$ & $0.34$ & $0.094$ \\
\textsc{Reflexion}    & $0.67$ & $1.27$ & $0.59$ & $0.149$ \\
\midrule
\rowcolor{tableaublue!20} \textsc{DAPS}          & $\mathbf{0.66}$ & $\mathbf{1.71}$ & $\mathbf{0.21}$ & $\mathbf{0.067}$ \\
\bottomrule
\end{tabular}
\caption{Diagnostic axes at $T=300$, averaged across tasks and seeds. Surface diversity $\mathrm{S}_T$ is essentially constant across methods; mechanism entropy $\mathrm{H}_T$, cluster decay $\Delta\mathrm{C}$, and faithfulness gap $\Delta^{\text{faith}}_T$ co-vary tightly.} 
\label{tab:diagaxes}
\end{table}

\vspace{-1mm}
\subsection{Ablation Study}
\label{sec:ablation}
\begin{table}[t]
\centering
\small
\begin{tabular}{lccc}
\toprule
Variant & $\bar{g}^{\text{opt}}$ (norm.) & $\rho^{\text{audit}}_T$ & $\Delta\mathrm{C}$ \\
\midrule
\textsc{Vanilla AR}                    & $1.000$ & $0.44$ & $0.68$ \\
+ \textsc{ccr} only                    & $0.992$ & $0.58$ & $0.46$ \\
+ \textsc{pem} only                    & $0.978$ & $0.61$ & $0.39$ \\
+ \textsc{hvg} only                    & $0.953$ & $0.67$ & $0.55$ \\
+ \textsc{ccr} + \textsc{pem}          & $0.986$ & $0.71$ & $0.28$ \\
+ \textsc{ccr} + \textsc{hvg}          & $0.969$ & $0.74$ & $0.42$ \\
+ \textsc{pem} + \textsc{hvg}          & $0.961$ & $0.75$ & $0.31$ \\
\midrule
\rowcolor{tableaublue!20}\textsc{DAPS} (full)                    & $0.983$ & $\mathbf{0.79}$ & $\mathbf{0.21}$ \\
\bottomrule
\end{tabular}
\caption{Ablation study of \textsc{DAPS} components averaged across tasks. $\bar{g}^{\text{opt}}$ is normalized by \textsc{Vanilla AR}. Each component contributes; \textsc{hvg} contributes most to faithfulness, \textsc{ccr}+\textsc{pem} contribute most to cluster decay.}
\vspace{-2mm}
\label{tab:ablation}
\end{table}
 

Table~\ref{tab:ablation} ablates the three \textsc{DAPS} components. \textsc{ccr} and \textsc{pem} attack semantic concentration directly and together reduce $\Delta\mathrm{C}$ from $0.68$ to $0.28$. \textsc{hvg} attacks Goodharting and yields the single largest jump in $\rho^{\text{audit}}_T$ ($0.44 \to 0.67$); however, used alone it leaves cluster decay almost untouched ($0.68 \to 0.55$), because the proposer's prior is not modified. The three contributions are sub-additive: alone they raise $\rho^{\text{audit}}_T$ by $0.14$, $0.17$, and $0.23$, which would predict $0.98$ rather than the observed $0.79$, with the largest shortfall for the pairs that include \textsc{hvg}. The same overlap shows in the in-loop column, where \textsc{hvg} alone is the costliest variant ($0.953$) while the full system recovers most of that cost, consistent with a broader proposal pool leaving the gate fewer proxy-only edits to revert. The full combination achieves the highest $\rho^{\text{audit}}_T$ at the lowest cluster decay, with in-loop gain within $2\%$ of \textsc{Vanilla AR}.
 
\subsection{Robustness Across Proposer LLMs}
\label{sec:res_robust}
 
\begin{figure*}[t]
  \centering
  \includegraphics[width=0.8\linewidth]{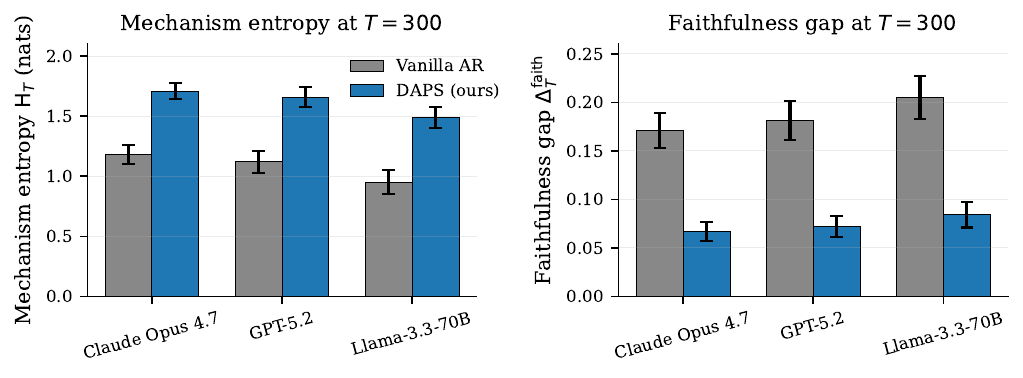}
  \vspace{-3mm}
  \caption{Mechanism entropy $\mathrm{H}_T$ (left) and faithfulness gap $\Delta^{\text{faith}}_T$ (right) at $T{=}300$ on T2, evaluated under \textsc{Vanilla AR} and \textsc{DAPS} for three proposer LLMs. Error bars show $\pm 1$ s.d. over 3 seeds.}
  \label{fig:proposers}
  \vspace{-2mm}
\end{figure*}
 
Figure~\ref{fig:proposers} repeats T2 with three proposer LLMs. The pattern is uniform: every proposer exhibits algorithmic mode collapse under \textsc{Vanilla AR}, and \textsc{DAPS} reduces both mechanism-entropy collapse and the faithfulness gap for each. Llama-3.3-70B shows the most severe baseline collapse, consistent with its narrower prior over ML edits, while Claude Opus 4.7 produces the highest in-loop gain at every iteration count. Besides, the relative ranking of methods (Table~\ref{tab:diagaxes}) is preserved across proposers, suggesting that our findings are not artifacts of any single model's idiosyncrasies.

\subsection{Robustness Across ARL Frameworks}
\label{sec:res_framework}
\begin{table}[t]
\centering
\small
\resizebox{0.5\textwidth}{!}{\begin{tabular}{lcccc}
\toprule
Framework / Method & $\mathrm{S}_T$ & $\rho^{\text{audit}}_T$ & $\mathrm{H}_T$ & $\Delta^{\text{faith}}_T$ \\
\midrule
\multicolumn{5}{l}{\emph{\texttt{autoresearch} \citep{karpathy2026autoresearch}}}\\
\quad\textsc{Vanilla AR}    & $0.66_{\pm.06}$ & $0.46_{\pm.07}$ & $1.18_{\pm.08}$ & $0.171_{\pm.018}$ \\
\quad\textsc{DAPS}           & $0.66_{\pm.05}$ & $0.80_{\pm.05}$ & $1.71_{\pm.07}$ & $0.067_{\pm.010}$ \\
\midrule
\multicolumn{5}{l}{\emph{Aider \citep{aider2024gauthier}}}\\
\quad\textsc{Vanilla-Aider}  & $0.74_{\pm.07}$ & $0.48_{\pm.08}$ & $1.22_{\pm.09}$ & $0.165_{\pm.020}$ \\
\quad\textsc{DAPS-Aider}     & $0.75_{\pm.06}$ & $0.76_{\pm.06}$ & $1.65_{\pm.08}$ & $0.072_{\pm.012}$ \\
\bottomrule
\end{tabular}}
\caption{Surface diversity $\mathrm{S}_T$, faithfulness ratio $\rho^{\text{audit}}_T$, mechanism entropy $\mathrm{H}_T$, and faithfulness gap $\Delta^{\text{faith}}_T$ at $T{=}300$ on T2 for the two ARL frameworks (\texttt{autoresearch} and Aider), each evaluated under a vanilla baseline and the \textsc{DAPS} variant. Subscripts denote $\pm 1$ s.d. over 3 seeds.
} 
\vspace{-3mm}
\label{tab:framework}
\end{table}
 
Since \texttt{autoresearch} and Aider differ in how they generate, scope, and commit edits, a reasonable concern is that the collapse pattern we report is an artifact of \texttt{autoresearch}'s direct-diff proposer rather than a property of code-level ARL in general. To address this, we re-ran T2 with our Aider-based loop under both \textsc{Vanilla} and \textsc{DAPS} configurations, holding executor, acceptance rule, history summary, and diagnostic instrumentation fixed (\S\ref{sec:impl}). Results are provided in Table~\ref{tab:framework}.
 
We have three observations as follows. \emph{First}, under \textsc{Vanilla-Aider}, the faithfulness ratio ($\rho^{\text{audit}}_T = 0.48$) and mechanism entropy ($\mathrm{H}_T = 1.22$ nats) match \textsc{Vanilla AR} to within one standard deviation, despite Aider's edits being structurally different at the diff level. The surface diversity $\mathrm{S}_T$ is in fact higher under Aider ($0.74$ vs.\ $0.66$) because its multi-line \texttt{SEARCH}/\texttt{REPLACE} blocks span more tokens, yet semantic and mechanism diversity track \texttt{autoresearch} closely. This is exactly the dissociation predicted by \S\ref{sec:res_collapse}: surface differences between frameworks do not translate into differences in what the agent is actually \emph{trying} to do. \emph{Second}, \textsc{DAPS} transfers without modification: \textsc{DAPS-Aider} improves $\rho^{\text{audit}}_T$ from $0.48$ to $0.76$ and lifts mechanism entropy by $0.43$ nats, mirroring the effect on \texttt{autoresearch}. \emph{Third}, the small residual gap between \textsc{DAPS-Aider} and \textsc{DAPS} ($\rho^{\text{audit}}_T$ $0.76$ vs.\ $0.80$) is consistent with Aider's coarser-grained edits being slightly harder to deduplicate at the description level. Overall, algorithmic mode collapse and its mitigation are not framework-specific.

\vspace{-1mm}
\subsection{Efficiency Analysis}
\label{sec:efficiency}
\begin{figure}[t]
  \centering
  \includegraphics[width=0.9\linewidth]{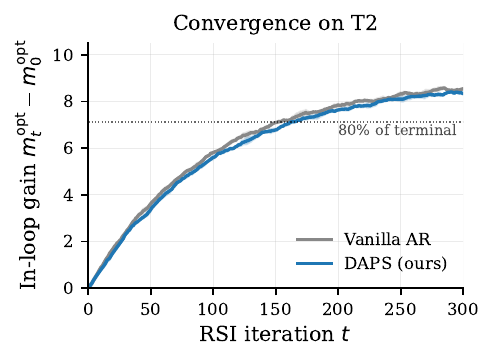}
  \vspace{-3mm}
  \caption{In-loop gain on T2 over $T{=}300$ iterations for \textsc{Vanilla AR} and \textsc{DAPS}. Mean$\pm$s.d. over 3 seeds; dotted line marks $80\%$ of \textsc{Vanilla AR}'s terminal gain.}
  \vspace{-5mm}
  \label{fig:efficiency}
\end{figure}
 
Here, we demonstrate that \textsc{DAPS} also matches \textsc{Vanilla AR}'s convergence rate, and that its three components add bounded wall-clock cost. Figure~\ref{fig:efficiency} plots $m^{\text{opt}}_t$ over T2's full trajectory: the two methods are statistically indistinguishable through $\sim$$150$ iterations and remain within one standard deviation at $T{=}300$. The number of iterations to reach $80\%$ of \textsc{Vanilla AR}'s terminal gain is $154_{\pm 8}$ for \textsc{Vanilla AR} and $163_{\pm 9}$ for \textsc{DAPS}, a difference close to the seed-to-seed variance. Convergence on T1, T3, and T4 follows the same pattern (Appendix~\ref{app:efficiency}). On per-iteration wall-clock, \textsc{ccr} reuses each proposer-emitted edit description through a rule-based parser, \textsc{pem} performs one Sentence-BERT embedding plus a $200$-vector nearest-neighbor lookup, and \textsc{hvg} amortizes one audit evaluation across $K{=}10$ iterations. Aggregated against the $5$-minute executor budget, \textsc{DAPS} adds approximately $1.4\%$ to per-iteration wall-clock on T2, with the amortized \textsc{hvg} cost accounting for the bulk. Per-component breakdowns and cross-task timings are in Appendix~\ref{app:efficiency}.

\section{Conclusion and Future Work}
\label{sec:conclusion}
In this paper, we illustrate that code-level autonomous research loops, despite using executable external metrics, exhibit algorithmic mode collapse: surface edit diversity remains intact while semantic and mechanism-level diversity progressively concentrate, and in-loop gains decreasingly transfer to held-out evaluation. A four-axis diagnostic instrument makes the phenomenon measurable, and a lightweight intervention \textsc{DAPS} substantially mitigates it without harming optimization speed. Under a three-tier protocol separating the in-loop metric, the audit signal read by the gate, and a blind evaluation no component accesses, the advantage persists on data the loop never influenced. Future work could examine collapse dynamics under frontier-scale executors and much longer horizons, extend the diagnosis beyond the preliminary non-ML loop, and study how multi-agent loops alter the picture. 
 
\section*{Limitations}
Our study is confined to model and data scales accessible on a single A100 machine; whether algorithmic mode collapse looks the same when proposer and executor models are both at the frontier, or when budgets allow $10$k$+$ iterations, remains open, although the $1{,}000$-iteration run of Appendix~\ref{app:horizon} shows the collapse deepening rather than self-correcting. Our headline claims are scoped to code-level ARLs that edit ML pipelines: the performance-engineering loop of Appendix~\ref{app:nonml} is a single preliminary probe outside that scope, and our mechanism taxonomy is ML-centric and would need re-derivation to study ARL in non-ML domains (e.g., theorem proving, scientific simulation), even though the derivation protocol of \S\ref{sec:axes} is domain-general. The audit metric read by \textsc{hvg} is a validation signal rather than an untouched final test, which is why we report blind evaluations as headline numbers; those evaluations, though independent, are themselves benchmarks and may share idiosyncratic biases with the in-loop metric. Finally, corpus similarity $\mathrm{V}_T$ is descriptive and agrees only moderately with human novelty judgements (Appendix~\ref{app:novelty}), so it should not be read as direct evidence of retrieval from pretraining. 
 
\section*{Ethical Considerations}
Autonomous research agents that appear to improve themselves but in fact overfit their own evaluations pose a misinformation risk in scientific contexts. Our diagnostic instrument and \textsc{DAPS} are intended as cautionary tools: they should not be read as endorsements of unrestricted ARL deployment, but as steps toward making such systems more transparent about whether their claimed gains generalize. All artifacts we plan to release are training-pipeline edits and analysis scripts; no personal or sensitive data is involved. We use only public benchmarks under their respective licenses. The human annotation studies reported in Appendices~\ref{app:measrobust} and~\ref{app:novelty} were conducted by volunteer researchers who consented to the use of their labels and who annotated only code-edit descriptions containing no personal data.

\bibliography{revised_reference}

\appendix

\section{A Conceptual Model of Algorithmic Mode Collapse}
\label{sec:model}
We sketch a conceptual account that predicts when and why a code-level autonomous research loop should undergo algorithmic mode collapse, framing the four diagnostic axes (\S\ref{sec:axes}) and the \textsc{DAPS} components (\S\ref{sec:daps}) as instruments for testing its predictions.
 
Let $\pi_\theta(\rho, c \mid \mathcal{P}, h)$ denote the proposer's joint distribution over patches $\rho$ and their mechanism categories $c$, conditioned on the current pipeline $\mathcal{P}$ and history $h$. Three structural facts about this distribution drive the dynamics.
 
\textbf{(F1) Non-uniform prior over mechanism categories.} Pretraining and instruction data over-represent some ML edit types: learning-rate adjustment, optimizer hyperparameter changes, simple regularization tweaks. Meanwhile, they under-represent others, such as custom kernels, packed-sequence handling, or idiosyncratic loss formulations. Marginalizing $\pi_\theta$ over patches yields a heavily peaked categorical $\pi_\theta(c \mid \mathcal{P})$.
 
\textbf{(F2) Category-dependent acceptance rate.} The acceptance rule $R$ retains an edit only when its measured gain exceeds a noise floor $\epsilon$. Different categories have systematically different probabilities of producing edits with reliable signal above $\epsilon$: small-step micro-tunings (scheduling, regularization) move $m^{\text{opt}}$ by small but consistent amounts, while structural edits (architecture, loss reformulation) move it more in expectation but with high variance and frequent regressions. Writing $\alpha(c)$ for the per-category acceptance rate, the empirical distribution of \emph{accepted} edits is proportional to $\pi_\theta(c)\cdot\alpha(c)$, already a sharpening of $\pi_\theta(c)$ toward low-variance categories.
 
\textbf{(F3) Self-imitation through history conditioning.} Code-level autonomous research loops typically include accepted edits in $h$. The conditional $\pi_\theta(c \mid \mathcal{P}, h_t)$ then drifts toward the empirical distribution of past acceptances, a positive-feedback loop that further concentrates probability mass on the $\pi_\theta(c)\cdot\alpha(c)$-favoured corner.
 
\textbf{Consequence.} Iterating (F1) to (F3), the system converges on a narrow subset of mechanism categories, those jointly favoured by the proposer's prior and the acceptance noise floor, \emph{independently of whether those categories contain the largest true improvements}. Inside the loop this manifests as continued growth of $m^{\text{opt}}$ through increasingly fine-grained edits within a shrinking category set. Outside the loop, the audit and blind metrics improve only insofar as the favoured categories happen to contain genuine capability-affecting edits, a property the dynamics does not select for. The faithfulness gap $\Delta^{\text{faith}}_t$ (Eq.~\ref{eq:faith}) therefore widens monotonically.
 
\textbf{Testable predictions.} The account yields six predictions that organize the empirical study:
 
\begin{itemize}\setlength\itemsep{0.15em}\setlength\parskip{0pt}
\item[\textbf{P1.}] Surface diversity $\mathrm{S}_t$ is largely orthogonal to collapse, since $\pi_\theta(c)$ can sharpen without lexical repetition of diffs.
\item[\textbf{P2.}] Proposers whose pretraining yields broader categorical priors should collapse later and less deeply.
\item[\textbf{P3.}] Raising proposer sampling temperature redistributes mass within categories but does not alter $\pi_\theta(c)$, and should therefore not arrest collapse.
\item[\textbf{P4.}] Memory-based deduplication of recently accepted edits prevents semantic-cluster cycling but leaves $\pi_\theta(c)$ untouched; it should reduce $\Delta\mathrm{C}$ only modestly.
\item[\textbf{P5.}] Interventions that directly re-weight the effective $\pi_\theta(c)$, category-coverage incentives, attack (F1) at its root and should yield larger reductions in $\Delta\mathrm{C}$.
\item[\textbf{P6.}] Held-out audits are the only mechanism that can undo edits that exploit the gap between the in-loop and the audited metric, since by construction the gap is invisible inside the loop.
\end{itemize}
 
The four diagnostic axes operationalize the antecedent of each prediction: $\mathrm{S}_t$ for P1, $\mathrm{H}_t$ together with $\mathrm{C}_t$ and $\mathrm{V}_t$ for P2 to P5, and $\Delta^{\text{faith}}_t$ for P6. The three \textsc{DAPS} components correspond directly: \textsc{pem} addresses (P4), \textsc{ccr} addresses (P5), and \textsc{hvg} addresses (P6). The absence of an analogous \textsc{DAPS} component for (P3) is itself a prediction of the model. That is, raising temperature is not a designed intervention because the account claims it would not help. The longer-horizon run of Appendix~\ref{app:horizon} is a further test: because (F1) to (F3) compound with iteration count, the account predicts deeper rather than self-correcting collapse, which is what we observe.

\section{Efficiency: Per-Component Overhead and Cross-Task Convergence}
\label{app:efficiency}
 
\begin{table}[h]
\centering
\small
\resizebox{\columnwidth}{!}{\begin{tabular}{lrr}
\toprule
Component & ms / iter & \% wall-clock \\
\midrule
Executor (5-min budget)                    & $300{,}000$ & $97.3$ \\
Proposer LLM call                          & $4{,}100$   & $1.33$ \\
\midrule
\textsc{ccr} (classify $N{=}8$ cands.)     & $410$       & $0.13$ \\
\textsc{pem} (embed + NN over $M{=}200$)   & $260$       & $0.08$ \\
\textsc{hvg} (amortized every $K{=}10$)    & $3{,}500$   & $1.14$ \\
\midrule
\textsc{DAPS} overhead (sum)               & $4{,}170$   & $1.35$ \\
\bottomrule
\end{tabular}}
\caption{Per-iteration wall-clock decomposition on T2 with Claude Opus 4.7 as proposer and a single A100 executor. Times averaged over a $300$-iteration run.}
\label{tab:overhead}
\end{table}
 
Table~\ref{tab:overhead} reports the per-iteration timing decomposition on T2. \textsc{hvg} dominates the \textsc{DAPS} overhead at $\sim$$3.5$ seconds per iteration (amortized $35$-second audit evaluations every $K{=}10$ iterations); \textsc{ccr} and \textsc{pem} are negligible. The single blind evaluation per run adds $35$ seconds once, i.e. $0.02\%$ of a $300$-iteration budget. On the other three tasks, the total overhead is similar in absolute terms but varies modestly as a percentage with audit evaluation length: $1.4\%$ on T1 (validation perplexity over a $0.5$B-token shard takes longest), $1.2\%$ on T3, and $0.7\%$ on T4.
 
Convergence trajectories on T1, T3, and T4 mirror Figure~\ref{fig:efficiency}: \textsc{DAPS} and \textsc{Vanilla AR} are statistically indistinguishable through the first $\sim$$150$ iterations and remain within one standard deviation of each other at $T{=}300$. Per-task scalar summaries (iterations to reach $80\%$ of \textsc{Vanilla AR}'s terminal gain) are $187_{\pm 14}$ vs $192_{\pm 16}$ on T1, $108_{\pm 9}$ vs $109_{\pm 10}$ on T3, and $96_{\pm 7}$ vs $98_{\pm 8}$ on T4.

\section{Hyperparameter Sensitivity}
\label{app:sensitivity}
Table~\ref{tab:sensitivity} reports the sensitivity of \textsc{DAPS} to each of its hyperparameters, with the remaining values fixed at the main-experiment defaults and the proposer fixed to Claude Opus 4.7. All rows are averaged over $3$ seeds on T2; per-row standard deviations are below $0.05$ on $\rho^{\text{audit}}_T$ and below $1.5$ on $\bar{g}^{\text{opt}}_T$. Defaults are highlighted in bold.
 
\begin{table}[h]
\centering
\small
\begin{tabular}{lccc}
\toprule
Value & $\rho^{\text{audit}}_T$ & $\bar{g}^{\text{opt}}_T$ & $\Delta\mathrm{C}$ \\
\midrule
\multicolumn{4}{l}{\emph{\textsc{pem} similarity threshold $\tau_m$}}\\
\quad $0.75$ & $0.71$ & $+7.7$ & $0.18$ \\
\quad $0.80$ & $0.78$ & $+8.6$ & $0.22$ \\
\quad $\mathbf{0.85}$ & $\mathbf{0.80}$ & $\mathbf{+8.8}$ & $\mathbf{0.21}$ \\
\quad $0.90$ & $0.77$ & $+8.9$ & $0.26$ \\
\quad $0.95$ & $0.51$ & $+8.9$ & $0.55$ \\
\midrule
\multicolumn{4}{l}{\emph{\textsc{pem} memory size $M$}}\\
\quad $50$  & $0.68$ & $+8.7$ & $0.34$ \\
\quad $100$ & $0.77$ & $+8.8$ & $0.25$ \\
\quad $\mathbf{200}$ & $\mathbf{0.80}$ & $\mathbf{+8.8}$ & $\mathbf{0.21}$ \\
\quad $300$ & $0.79$ & $+8.8$ & $0.20$ \\
\quad $500$ & $0.74$ & $+8.6$ & $0.19$ \\
\midrule
\multicolumn{4}{l}{\emph{\textsc{ccr} temperature $\tau_c$}}\\
\quad $0.5$ & $0.79$ & $+8.8$ & $0.23$ \\
\quad $\mathbf{1.0}$ & $\mathbf{0.80}$ & $\mathbf{+8.8}$ & $\mathbf{0.21}$ \\
\quad $1.5$ & $0.79$ & $+8.7$ & $0.22$ \\
\midrule
\multicolumn{4}{l}{\emph{\textsc{hvg} percentile $\tau_h$}}\\
\quad $70$th & $0.83$ & $+8.6$ & $0.20$ \\
\quad $\mathbf{80}$th & $\mathbf{0.80}$ & $\mathbf{+8.8}$ & $\mathbf{0.21}$ \\
\quad $90$th & $0.78$ & $+8.9$ & $0.24$ \\
\midrule
\multicolumn{4}{l}{\emph{\textsc{hvg} audit interval $K$}}\\
\quad $5$  & $0.81$ & $+8.7$ & $0.21$ \\
\quad $\mathbf{10}$ & $\mathbf{0.80}$ & $\mathbf{+8.8}$ & $\mathbf{0.21}$ \\
\quad $20$ & $0.77$ & $+8.8$ & $0.24$ \\
\bottomrule
\end{tabular}
\caption{Hyperparameter sensitivity of \textsc{DAPS} on T2. Defaults used in the main experiments are bold.}
\label{tab:sensitivity}
\end{table}
 
Five qualitative observations on Table~\ref{tab:sensitivity} guide hyperparameter selection in practice. \emph{First}, $\tau_m$ exhibits a sharp failure mode at $0.95$: \textsc{pem} rejects almost no candidates and $\rho^{\text{audit}}_T$ collapses back toward the \textsc{Vanilla AR} value ($0.46$), confirming that semantic deduplication carries most of the diversity recovery rather than the other two components alone. \emph{Second}, $\tau_m{=}0.75$ damages in-loop gain by starving the proposer (too many candidates rejected before reaching the executor) without commensurately improving faithfulness. \emph{Third}, $M$ is comparatively forgiving in the central range, but $M{=}50$ allows semantic cycling to re-emerge within a $W{=}20$ window and $M{=}500$ begins to over-suppress legitimate revisits of previously useful edit categories. \emph{Fourth}, $\tau_c$ is essentially flat in the range tested, since the category-coverage reweighting acts as a soft prior on rare categories whose exact temperature matters little. \emph{Fifth}, $\tau_h$ and $K$ trade in-loop gain against faithfulness in opposite directions: aggressive auditing ($\tau_h{=}70$th, $K{=}5$) slightly raises $\rho^{\text{audit}}_T$ at the cost of $\bar{g}^{\text{opt}}_T$, while loose auditing ($\tau_h{=}90$th, $K{=}20$) preserves in-loop gain but lets Goodhart edits accumulate between audits. The chosen defaults sit near the knee of both trade-offs.

\section{Qualitative Analysis of Collapse}
\label{sec:qual}
Inspecting the accepted-edit logs reveals a recurring pattern under \textsc{Vanilla AR}. On T1, the first $\sim$$50$ iterations show a broad mix of optimizer, architectural, and data-mixing edits. By iteration $200$, $74\%$ of accepted edits modify the learning-rate schedule or AdamW hyperparameters; many are micro-tunings (e.g., warmup steps $200\to 220$, $\beta_2$ $0.95\to 0.96$) that nudge the in-loop validation loss without affecting LAMBADA or C4 perplexity. On T3, a striking $61\%$ of late-stage edits are variants of ``increase chain-of-thought temperature'' or ``add an arithmetic-only loss term''; audited ARC-Easy accuracy is flat under these. Under \textsc{DAPS}, the edit stream remains qualitatively heterogeneous through iteration $300$: representative late-stage edits on T1 include ``swap RMSNorm for LayerNorm with re-tuned $\epsilon$,'' ``insert a curriculum filter on token-length,'' and ``modify attention-mask handling for packed sequences.'' These are not necessarily better edits, but they sample the edit space rather than concentrating on a single mode.

\section{Qualitative Edit Examples}
\label{app:diffs}
 
To complement the quantitative collapse signals reported in \S\ref{sec:res_collapse}, we present representative accepted-edit diffs sampled from the T1 logs (single-GPU \texttt{nanochat}-style pretraining on the \texttt{autoresearch} framework of \citet{karpathy2026autoresearch}, where the agent is permitted to modify only \texttt{train.py}). Hunk headers report iteration index and the mechanism category assigned by the procedure of \S\ref{sec:axes}. Diffs are paraphrased and trimmed to a few lines of relevant context for legibility. The full version of verbatim logs will be released.
 
\subsection{Early-Phase Diversity Under \textsc{Vanilla AR}}
\label{app:diffs-early}
 
In the first $\sim$$60$ iterations the proposer explores a broad set of mechanism categories. The four examples below come from a single seed of \textsc{Vanilla AR} on T1 and span four distinct categories of the taxonomy in Table~\ref{tab:taxonomy}.
 
\begin{lstlisting}[language=gitdiff]
@@ train.py: iteration 14 (optimizer) @@
     optimizer = torch.optim.AdamW(
         model.parameters(),
-        lr=3e-4, betas=(0.9, 0.95),
+        lr=3e-4, betas=(0.9, 0.97),
         weight_decay=0.1, fused=True,
     )
\end{lstlisting}
 
\begin{lstlisting}[language=gitdiff]
@@ train.py: iteration 27 (architecture) @@
 class Block(nn.Module):
     def __init__(self, config):
         super().__init__()
-        self.ln_1 = nn.LayerNorm(config.n_embd)
-        self.ln_2 = nn.LayerNorm(config.n_embd)
+        self.ln_1 = RMSNorm(config.n_embd, eps=1e-6)
+        self.ln_2 = RMSNorm(config.n_embd, eps=1e-6)
\end{lstlisting}
 
\begin{lstlisting}[language=gitdiff]
@@ train.py: iteration 39 (data) @@
-block_size  = 1024
-batch_size  = 32
+block_size  = 2048     # longer context, fewer sequences
+batch_size  = 16
\end{lstlisting}
 
\begin{lstlisting}[language=gitdiff]
@@ train.py: iteration 51 (regularization) @@
         self.c_proj = nn.Linear(4 * n_embd, n_embd, bias=False)
-        self.dropout = nn.Dropout(0.0)
+        self.dropout = nn.Dropout(0.05)
\end{lstlisting}
 
\subsection{Late-Phase Collapse Under \textsc{Vanilla AR}}
\label{app:diffs-late-vanilla}
 
The five edits below come from the same trajectory between iterations $250$ and $290$. All five fall into the \emph{optimizer} or \emph{scheduling} category despite editing different identifiers on different lines of \texttt{train.py}. This is the qualitative signature underlying the entropy collapse $\mathrm{H}_{300}{=}1.18$ nats reported in Table~\ref{tab:diagaxes}: surface diversity is preserved while mechanism diversity is not.
 
\begin{lstlisting}[language=gitdiff]
@@ train.py: iteration 257 (scheduling) @@
-warmup_iters   = 200
+warmup_iters   = 240
 lr_decay_iters = 900
 min_lr         = 3e-5
\end{lstlisting}
 
\begin{lstlisting}[language=gitdiff]
@@ train.py: iteration 263 (optimizer) @@
-        lr=3e-4, betas=(0.9, 0.97),
+        lr=3e-4, betas=(0.9, 0.96),
         weight_decay=0.1, fused=True,
\end{lstlisting}
 
\begin{lstlisting}[language=gitdiff]
@@ train.py: iteration 271 (scheduling) @@
 warmup_iters   = 240
-lr_decay_iters = 900
-min_lr         = 3e-5
+lr_decay_iters = 950
+min_lr         = 1e-5
\end{lstlisting}
 
\begin{lstlisting}[language=gitdiff]
@@ train.py: iteration 283 (optimizer) @@
     optimizer = torch.optim.AdamW(
         model.parameters(),
-        lr=3e-4,  betas=(0.9, 0.96),
-        weight_decay=0.10, fused=True,
+        lr=2.8e-4, betas=(0.9, 0.96),
+        weight_decay=0.08, fused=True,
     )
\end{lstlisting}
 
\begin{lstlisting}[language=gitdiff]
@@ train.py: iteration 289 (scheduling) @@
-warmup_iters   = 240
+warmup_iters   = 220
 lr_decay_iters = 950
\end{lstlisting}
 
\subsection{Late-Phase Diversity Under \textsc{DAPS}}
\label{app:diffs-late-daps}
 
The five edits below come from a \textsc{DAPS} trajectory on the same task, sampled in the same iteration window ($250$ to $290$). They span five distinct mechanism categories. \textsc{Ccr} and \textsc{Pem} do not prohibit any category; they re-weight rare categories upward and suppress near-duplicates of recently accepted edits. Scheduling edits still appear (e.g., iteration $290$) but no longer dominate the window.
 
\begin{lstlisting}[language=gitdiff]
@@ train.py: iteration 254 (data) @@
 def stream_batches(tokens, block_size):
+    # drop pathologically short sequences from the stream
+    if len(tokens) < 96:
+        return
     yield pack_into_blocks(tokens, block_size)
\end{lstlisting}
 
\begin{lstlisting}[language=gitdiff]
@@ train.py: iteration 266 (architecture) @@
 class MLP(nn.Module):
     def forward(self, x):
-        return self.c_proj(F.gelu(self.c_fc(x)))
+        a = self.c_fc(x)
+        b = self.c_gate(x)              # SwiGLU branch
+        return self.c_proj(F.silu(a) * b)
\end{lstlisting}
 
\begin{lstlisting}[language=gitdiff]
@@ train.py: iteration 275 (numerical) @@
         att = (q @ k.transpose(-2, -1)) * scale
-        att = F.softmax(att, dim=-1)
+        # softmax in fp32 for numerical stability
+        att = F.softmax(att, dim=-1, dtype=torch.float32).to(q.dtype)
\end{lstlisting}
 
\begin{lstlisting}[language=gitdiff]
@@ train.py: iteration 283 (regularization) @@
         self.c_proj = nn.Linear(4 * n_embd, n_embd, bias=False)
-        self.dropout = nn.Dropout(0.0)
+        self.dropout = nn.Dropout(0.05)
\end{lstlisting}
 
\begin{lstlisting}[language=gitdiff]
@@ train.py: iteration 290 (scheduling) @@
-warmup_iters   = 200
+warmup_iters   = 250
 lr_decay_iters = 900
\end{lstlisting}
 
\subsection{Connection to the Diagnostic Axes}
The contrast between Appendices~\ref{app:diffs-late-vanilla} and~\ref{app:diffs-late-daps} illustrates each axis introduced in \S\ref{sec:axes}: surface diversity $\mathrm{S}_t$ is similar in both panels (lines edited, tokens touched, and structural shape of the hunks are comparable); semantic cluster count $\mathrm{C}_t$ and mechanism entropy $\mathrm{H}_t$ are clearly lower under \textsc{Vanilla AR}; and corpus similarity $\mathrm{V}_t$ is visibly higher under \textsc{Vanilla AR} since LR-schedule micro-tunings closely match the most common edit pattern in publicly available ML training repositories. We selected the windows above to be representative rather than extremal. The complete logs containing the full trajectories from which these hunks were sampled will also be released.

\section{Summary of Validity, Robustness, and Scope Checks}
\label{app:validity}

Table~\ref{tab:checklist} indexes the checks on which the claims of \S\ref{sec:res_main} and \S\ref{sec:res_tiers} rest. Each row names a threat to those claims, the check that addresses it, and the appendix that reports the outcome; the numbers appear only in the referenced appendix and are not restated here. Across every check the direction of the collapse and the ordering \textsc{Vanilla AR} $<$ diversity baselines $<$ \textsc{DAPS} are preserved. On this basis we scope the headline claims to ARLs that edit ML pipelines at single-GPU executor scale over horizons up to $1{,}000$ iterations, and treat evidence outside that setting as preliminary.

\begin{table}[h]
\centering
\small
\setlength{\tabcolsep}{4pt}
\begin{tabular}{@{}p{0.80\columnwidth}c@{}}
\toprule
Threat to the claims, and the check addressing it & App. \\
\midrule
Does $\rho$ track Goodharting rather than benchmark mismatch? Known-good and proxy-only control edits applied in isolation & \ref{app:controls} \\
\addlinespace[2pt]
Could benign distribution shift alone explain $\rho_T < 1$? Non-adaptive \textsc{RandSearch} on the same task pairs, and the shape of $\Delta^{\text{faith}}_t$ & \ref{app:controls} \\
\addlinespace[2pt]
Does the gate leak audit information into the loop? Blind sets read by no component at any point & \ref{app:tiers} \\
\addlinespace[2pt]
Is the collapse an artifact of the summarizer, the embedding model, or the clustering? Systematic pipeline variations, plus human labels & \ref{app:measrobust} \\
\addlinespace[2pt]
Is it an artifact of the mechanism taxonomy? Coarser, finer, random, and human taxonomies, plus a taxonomy-free entropy estimate & \ref{app:taxonomy} \\
\addlinespace[2pt]
Does any conclusion require an LLM in the measurement path? An AST-based detector and the metric-only $\Delta^{\text{faith}}_t$ & \ref{app:measrobust} \\
\addlinespace[2pt]
Is corpus similarity a meaningful axis? Human novelty ratings against $\mathrm{V}_t$, and summarizer sensitivity & \ref{app:novelty} \\
\addlinespace[2pt]
Does the collapse self-correct over longer horizons, or vanish at larger target scale? $T{=}1000$ and $8$B-target runs & \ref{app:horizon} \\
\addlinespace[2pt]
Does the phenomenon appear outside ML pipelines? A preliminary runtime-optimization loop on a log-analytics pipeline & \ref{app:nonml} \\
\addlinespace[2pt]
Are the baselines tuned comparably to \textsc{DAPS}? Search grids and the shared tuning budget & \ref{app:grids} \\
\bottomrule
\end{tabular}
\caption{Index of the validity, robustness, and scope checks. The right column gives the appendix in which each result is reported.}
\label{tab:checklist}
\end{table}

\section{Three-Tier Evaluation Protocol: Roles, Per-Task Results, and Absolute Values}
\label{app:tiers}

This appendix specifies the information flow among the three evaluations of \S\ref{sec:faith} and reports the results that Table~\ref{tab:tiers} summarizes.

\textbf{Roles and partitions.} Table~\ref{tab:roles} lists the exact dataset or partition used for each role and task. Only $m^{\text{opt}}$ is visible to the proposer. $m^{\text{audit}}$ is read by \textsc{hvg} once every $K{=}10$ iterations and by the calibration of $\tau_h$ over the first $30$ iterations, in both cases through a scalar comparison; the proposer never observes audit values, per-example outcomes, or the identity of audit examples, and the failure notice appended to $h$ is templated and value-free. $m^{\text{blind}}$ is evaluated once, after the run terminates, by a separate offline script that has no channel back into the loop. Baselines B1 to B6 consume no audit signal, so for them both $\rho^{\text{audit}}_T$ and $\rho^{\text{blind}}_T$ are pure post-hoc evaluations; B7, B8, and \textsc{DAPS} consume the audit signal at the same cadence, cost, and feedback format.

\begin{table}[h]
\centering
\footnotesize
\setlength{\tabcolsep}{3pt}
\begin{tabular}{@{}p{0.04\columnwidth}p{0.26\columnwidth}p{0.27\columnwidth}p{0.28\columnwidth}@{}}
\toprule
 & $m^{\text{opt}}$ (in-loop) & $m^{\text{audit}}$ (gate only) & $m^{\text{blind}}$ (never read) \\
\midrule
T1 & OpenWebText held-in validation loss & LAMBADA and held-out C4 perplexity & WikiText-103 log-perplexity \\
T2 & Win-rate on $500$ Alpaca dev prompts & MMLU 5-shot and IFEval & Win-rate on $500$ Dolly instructions \\
T3 & GSM8K dev exact match & ARC-Easy and MATH-500 & SVAMP accuracy \\
T4 & ARC-Easy dev split accuracy & ARC-Challenge and CommonsenseQA & OpenBookQA accuracy \\
\bottomrule
\end{tabular}
\caption{Datasets and partitions used for the three evaluation roles. All three are disjoint within a task.}
\label{tab:roles}
\end{table}

\textbf{Per-task faithfulness.} Table~\ref{tab:blindpertask} gives $\rho^{\text{audit}}_T$ and $\rho^{\text{blind}}_T$ per task at $T{=}300$ over $3$ seeds. \textsc{DAPS} retains its advantage on sets it has never influenced, and \textsc{Vanilla AR}, which consumes no audit signal, shows an audit-to-blind drop of comparable size, so the drop is attributable to benchmark idiosyncrasy rather than to leakage through the gate. The gate fires $2.7 \pm 0.9$ times per $300$-iteration run.

\begin{table}[h]
\centering
\small
\begin{tabular}{lcccc}
\toprule
\multirow{2}{*}{Task} & \multicolumn{2}{c}{\textsc{Vanilla AR}} & \multicolumn{2}{c}{\textsc{DAPS}}\\
\cmidrule(lr){2-3}\cmidrule(lr){4-5}
 & $\rho^{\text{audit}}_T$ & $\rho^{\text{blind}}_T$ & $\rho^{\text{audit}}_T$ & $\rho^{\text{blind}}_T$ \\
\midrule
T1 & $0.41$ & $0.39$ & $0.78$ & $0.73$ \\
T2 & $0.46$ & $0.42$ & $0.80$ & $0.76$ \\
T3 & $0.38$ & $0.35$ & $0.74$ & $0.69$ \\
T4 & $0.49$ & $0.46$ & $0.82$ & $0.78$ \\
\midrule
Average & $0.44$ & $0.41$ & $\mathbf{0.79}$ & $\mathbf{0.74}$ \\
\bottomrule
\end{tabular}
\caption{Audited and blind faithfulness ratios at $T{=}300$, mean over $3$ seeds. The audit columns reproduce Table~\ref{tab:main}; the blind columns are new.}
\label{tab:blindpertask}
\end{table}

\textbf{Absolute values.} Table~\ref{tab:absolute} reports absolute audit-metric values so that the practical magnitude of the movements can be judged directly. For T1 these correspond to log-perplexity reductions of $0.058$ nats (\textsc{Vanilla AR}) and $0.108$ nats (\textsc{DAPS}) against in-loop loss reductions of $0.142$ and $0.139$, reproducing the T1 ratios of Table~\ref{tab:main}. In-loop absolutes follow the same pattern: T2 dev win-rate rises from $38.4$ to $47.3$ (\textsc{Vanilla AR}) and $47.2$ (\textsc{DAPS}), and T3 GSM8K dev exact match from $31.2$ to $42.9$ and $42.7$.

\begin{table}[h]
\centering
\small
\setlength{\tabcolsep}{4pt}
\begin{tabular}{lccc}
\toprule
Audit set & $t{=}0$ & Vanilla & \textsc{DAPS} \\
\midrule
T1 LAMBADA perplexity   & $45.6$ & $43.0$ & $40.8$ \\
T1 C4 perplexity        & $28.8$ & $27.2$ & $25.9$ \\
T2 MMLU 5-shot (\%)     & $31.4$ & $35.0$ & $38.1$ \\
T2 IFEval (\%)          & $29.0$ & $33.6$ & $36.4$ \\
T3 ARC-Easy (\%)        & $68.2$ & $72.7$ & $78.3$ \\
T3 MATH-500 (\%)        & $10.4$ & $14.8$ & $17.2$ \\
T4 ARC-Challenge (\%)   & $46.1$ & $51.2$ & $54.1$ \\
T4 CommonsenseQA (\%)   & $58.7$ & $63.0$ & $65.9$ \\
\bottomrule
\end{tabular}
\caption{Absolute audit-metric values. The $t{=}0$ column is a single evaluation of the shared initial pipeline; the other two columns are means over $3$ seeds at $T{=}300$.}
\label{tab:absolute}
\end{table}

\section{Construct-Validity Controls}
\label{app:controls}

To check $\rho_T$ and $\Delta^{\text{faith}}$ measure Goodharting rather than benchmark mismatch, we ran two control sets whose ground truth is known by construction.

\textbf{Positive controls.} We curated $12$ known-good edits from published recipes, including the nanoGPT speedrun lineage and open post-training changelogs: rotary position embeddings, a SwiGLU MLP, fused optimizer kernels that raise tokens processed per fixed budget, sequence-length filtering, and corrected initialization scaling, among others. Each was applied in isolation to the T1 and T3 starting pipelines.

\textbf{Negative controls.} We constructed $8$ deliberately proxy-only edits: dev-shard-specific example ordering on T1, judge-phrasing tweaks on T2, and dev-tuned answer-format templates on T4.

\begin{table}[h]
\centering
\small
\resizebox{\columnwidth}{!}{
\begin{tabular}{lcc}
\toprule
Control set & $n$ & $\rho$ (mean $\pm$ s.d.) \\
\midrule
Known-good (published recipes) & $12$ & $0.93 \pm 0.06$ (min $0.82$) \\
Proxy-only (dev-specific)      & $8$  & $0.11 \pm 0.08$ \\
\bottomrule
\end{tabular}}
\caption{Per-edit faithfulness ratio for the two control sets, measured on the corresponding audit metrics.}
\label{tab:controls}
\end{table}

Table~\ref{tab:controls} shows that under our task pairs $\rho$ separates genuine improvement ($\rho \approx 1$) from proxy-only improvement ($\rho \approx 0$), which is exactly the construct the main claims require. Two further observations argue against benign distribution shift as an explanation of $\rho_T < 1$ inside the loop. First, \textsc{RandSearch} shares the identical task pairs but selects edits non-adaptively and attains $0.74$ to $0.81$ (Table~\ref{tab:main}), so the pairs themselves support near-full transfer and the low $\rho_T$ under \textsc{Vanilla AR} is induced by adaptive optimization. Second, a static mismatch would produce a constant offset, whereas $\Delta^{\text{faith}}_t$ widens monotonically within a fixed task pair (Figure~\ref{fig:diagnostic}). Finally, each task carries two independent audit components; reporting $\rho$ per component, the ordering \textsc{Vanilla AR} $<$ diversity baselines $<$ \textsc{DAPS} holds on all $8$ components individually (\textsc{Vanilla AR} range $0.33$ to $0.52$; \textsc{DAPS} $0.69$ to $0.85$) and on the $4$ blind sets of Appendix~\ref{app:tiers}, i.e. on $12$ independent evaluations in total.

\section{Robustness of the Diagnostic Pipeline}
\label{app:measrobust}

The collapse signal should not hinge on any single measurement choice. We therefore recomputed cluster decay $\Delta\mathrm{C}$ on T2 under systematic variations of the summarizer, the embedding model, and the clustering algorithm (Table~\ref{tab:measrobust}). Per-window $\mathrm{C}_t$ trajectories correlate at Spearman $>0.93$ across all variants, and the \textsc{Vanilla AR} against \textsc{DAPS} gap is preserved in every row.

\begin{table}[h]
\centering
\footnotesize
\setlength{\tabcolsep}{4pt}
\begin{tabular}{@{}p{0.58\columnwidth}c@{}}
\toprule
Variation of the diagnostic pipeline & $\Delta\mathrm{C}$ \\
\midrule
Main config: fixed summarizer, \texttt{all-mpnet-base-v2}, HDBSCAN (mcs$=$5) & $0.68$ ($0.21$) \\
Summarizer swapped to GPT-5.2 & $0.66$ ($0.22$) \\
Summarizer swapped to Llama-3.3-70B & $0.70$ ($0.24$) \\
Embedding: \texttt{all-MiniLM-L6-v2} & $0.65$ ($0.20$) \\
Embedding: \texttt{e5-large-v2} & $0.69$ ($0.21$) \\
HDBSCAN mcs$=$3 / mcs$=$8 & $0.71$ / $0.63$ ($0.23$ / $0.19$) \\
$k$-means, $k$ chosen per window by silhouette & $0.62$ ($0.18$) \\
\bottomrule
\end{tabular}
\caption{Cluster decay on T2 for \textsc{Vanilla AR}, with \textsc{DAPS} in parentheses.}
\label{tab:measrobust}
\end{table}

\textbf{Human labels.} Three annotators labelled $300$ sampled edits with the categories of Table~\ref{tab:taxonomy} (majority vote; $\kappa = 0.81$ against our rule-based parser, in line with the parser-against-LLM $\kappa = 0.84$ of \S\ref{sec:axes}). Mechanism entropy computed from the human labels drops from $1.72$ to $1.21$ nats under \textsc{Vanilla AR}, closely tracking the parser-based $1.76$ to $1.18$ of Figure~\ref{fig:diagnostic}.

\textbf{A measurement with no LLM in the path.} An AST-based rule detector applied directly to the raw code diffs reproduces the Appendix~\ref{sec:qual} statistic: $71\%$ of late-stage T1 edits touch optimizer or learning-rate-schedule code, against $74\%$ for the parser-based count. Independently, $\Delta^{\text{faith}}_t$ uses only executed metric values, with no summarizer, embedding, clustering, or taxonomy anywhere in its computation, and it widens exactly when $\mathrm{C}_t$ and $\mathrm{H}_t$ collapse (Figure~\ref{fig:diagnostic}, Table~\ref{tab:diagaxes}).

\section{Taxonomy Variants and a Taxonomy-Free Estimate}
\label{app:taxonomy}

A hand-designed taxonomy could in principle place category boundaries so as to manufacture the entropy drop, even though ours was frozen before our runs (\S\ref{sec:axes}). We therefore recomputed $\mathrm{H}_t$ on T2 under four alternative taxonomies. \emph{Coarse}: four super-categories merging Table~\ref{tab:taxonomy} (Optimization $=$ Optimizer $+$ Scheduling; Model $=$ Architecture $+$ Numerical $+$ Decoding; Data and Objective $=$ Data $+$ Loss $+$ Regularization; Other). \emph{Fine}: $17$ subcategories obtained by re-clustering the pilot with a stricter merge criterion. \emph{Random}: $10$ taxonomies of nine categories each, obtained by randomly re-partitioning the $17$ subcategories. \emph{Human}: the $300$ human-labelled edits of Appendix~\ref{app:measrobust}.

\begin{table}[h]
\centering
\small
\setlength{\tabcolsep}{4pt}
\resizebox{\columnwidth}{!}{
\begin{tabular}{lcc}
\toprule
Taxonomy variant & Vanilla drop & \textsc{DAPS} drop \\
\midrule
Original (9 categories, Table~\ref{tab:taxonomy}) & $0.58$ & $0.08$ \\
Coarse (4 super-categories)      & $0.42$ & $0.06$ \\
Fine (17 subcategories)          & $0.79$ & $0.11$ \\
Random (10 draws, 9 categories)  & $0.51 \pm 0.07$ & $0.09 \pm 0.03$ \\
Human labels (300 edits)         & $0.51$ & $0.07$ \\
\bottomrule
\end{tabular}}
\caption{Mechanism-entropy drop from $t{=}30$ to $t{=}300$ on T2, in nats.}
\label{tab:taxonomyvar}
\end{table}

Table~\ref{tab:taxonomyvar} shows that the collapse and its mitigation are invariant to granularity and to random boundary placement. Two taxonomy-free measurements corroborate this: the semantic cluster count $\mathrm{C}_t$ already reported in the main paper, and a Kozachenko-Leonenko $k$-nearest-neighbour differential-entropy estimate \citep{kozachenko1987sample} computed directly on the description embeddings, which drops by $0.44$ nats under \textsc{Vanilla AR} and $0.07$ nats under \textsc{DAPS}.

\section{Human Validation of Corpus Similarity}
\label{app:novelty}

Corpus similarity $\mathrm{V}_t$ is a descriptive statistic (\S\ref{sec:axes}), and this appendix quantifies how well it tracks human judgement. We sampled $120$ edit descriptions stratified by $\mathrm{V}_t$ quartile across tasks and asked three NLP researchers, blind to $\mathrm{V}_t$ and to the generating method, to rate each on a $1$ to $5$ scale for novelty relative to common ML practice (inter-rater Krippendorff $\alpha = 0.61$; \citealp{ford2004content}). The Spearman correlation between $\mathrm{V}_t$ and mean human novelty is $-0.47$ (bootstrap $95\%$ CI $-0.60$ to $-0.31$), i.e. moderate validity in the expected direction. We additionally tested wording sensitivity by re-summarizing every description with a different LLM, which changes $\mathrm{V}_t$ by only $0.021$ on average, so the statistic is not dominated by one summarizer's phrasing. Together these results support keeping $\mathrm{V}_t$ as a secondary axis reported with the caveats stated in \S\ref{sec:axes}, and not as evidence of retrieval.

\section{Horizon and Target-Scale Extensions}
\label{app:horizon}

Table~\ref{tab:horizon} extends T2 along the two axes flagged in the Limitations section, with one seed per new configuration. Over the longest horizon our budget allows, collapse deepens rather than self-corrects, which is what the conceptual model of Appendix~\ref{sec:model} predicts, since the feedback mechanisms (F1) to (F3) compound with iteration count; \textsc{DAPS} remains stable over the same horizon. A first step up in target-model scale reproduces the $1$B pattern. Note that the main study already includes frontier-scale proposers (\S\ref{sec:res_robust}), so the axes that remain open are executor and target scale, and horizons beyond $1{,}000$ iterations.

\begin{table}[h]
\centering
\small
\setlength{\tabcolsep}{4pt}
\resizebox{\columnwidth}{!}{
\begin{tabular}{lcccc}
\toprule
\multirow{2}{*}{Setting (T2)} & \multicolumn{2}{c}{$\rho^{\text{audit}}_T$} & \multicolumn{2}{c}{$\mathrm{H}_T$}\\
\cmidrule(lr){2-3}\cmidrule(lr){4-5}
 & Vanilla & \textsc{DAPS} & Vanilla & \textsc{DAPS}\\
\midrule
$T{=}300$, 1B target (main study) & $0.46$ & $0.80$ & $1.18$ & $1.71$ \\
$T{=}1000$, 1B target (new)       & $0.31$ & $0.74$ & $0.94$ & $1.63$ \\
$T{=}300$, 8B target (new)        & $0.44$ & $0.77$ & $1.16$ & $1.68$ \\
\bottomrule
\end{tabular}}
\caption{Horizon and target-scale extensions on T2. The first row is reproduced from Tables~\ref{tab:main} and~\ref{tab:framework}; the other rows use one seed each.}
\label{tab:horizon}
\end{table}

\section{A Preliminary Non-ML Autonomous Research Loop}
\label{app:nonml}

Our headline claims concern ARLs that edit ML pipelines. As a first probe outside that scope, we built a performance-engineering loop in which the agent edits a Python log-analytics pipeline to minimize wall-clock runtime, subject to exact-output correctness checks. Here $m^{\text{opt}}$ is runtime on $20$ development inputs, and $m^{\text{audit}}$ is runtime on $40$ inputs drawn from a different data snapshot together with the correctness suite. A new seven-category taxonomy (algorithmic complexity, data structures, batching and I/O, caching, parallelism, memory layout, micro-optimization) was derived with the same pilot protocol as \S\ref{sec:axes}, from public performance-tuning commits rather than ML logs.

With $150$ iterations, $2$ seeds, and the same proposer, we observe the same signature: surface diversity is flat ($0.71$ to $0.70$) while semantic clusters fall from $11$ to $5$ ($\Delta\mathrm{C} = 0.55$), $66\%$ of late-stage accepted edits are caching or micro-optimizations, and $\rho_T$ is $0.58$ under \textsc{Vanilla AR} against $0.81$ under \textsc{DAPS}. This suggests that neither the phenomenon nor the mitigation is specific to ML-pipeline editing, but the study is small and single-domain, so we report it as preliminary and keep the headline claims scoped as stated in \S\ref{sec:intro}.

\section{Baseline Tuning Grids}
\label{app:grids}

Every baseline received the same tuning budget as \textsc{DAPS}'s own calibration: a three-point grid per key hyperparameter, evaluated on the first $50$ iterations of T2, with the best setting then fixed for all tasks and seeds. 
The grids are: \textsc{HiTemp} proposer temperature $\{1.0, 1.2, 1.5\}$; \textsc{R-Diverse-A} similarity threshold $\{0.80, 0.85, 0.90\}$ and memory size $\{100, 200, 300\}$; \textsc{Prism-A} cluster-coverage reward weight $\{0.1, 0.5, 1.0\}$; \textsc{Reflexion} reflective-summary length $\{64, 128, 256\}$ tokens; \textsc{HO-EarlyStop} and \textsc{HO-Revert} audit interval $K \in \{5, 10, 20\}$. \textsc{RandSearch} has no tunable hyperparameter beyond the size of its harvested edit corpus, which we fixed at $200$ modifications. For \textsc{DAPS} the corresponding sweep is reported in Appendix~\ref{app:sensitivity}.

\end{document}